\pdfoutput=1

\documentclass[authoryear]{elsarticle}

\usepackage{IEK10} 
\usepackage{natbib}
\usepackage[section]{placeins}
\usepackage{comment}

\def\IEK10{
  Institute of Climate and Energy Systems,
  Energy Systems Engineering (ICE-1),
  Forschungszentrum J\"ulich GmbH,
  J\"ulich 52425,
  Germany
}
\def\RWTH{
  RWTH Aachen University,
  Aachen 52062,
  Germany
}
\def\JARA{
  JARA-ENERGY,
  J{\"u}lich 52425,
  Germany
}
\def\SVT{
  RWTH Aachen University,
  Process Systems Engineering (AVT.SVT),
  Aachen 52074,
  Germany
}
\def\FVT{
  RWTH Aachen University,
  Fluid Process Engineering (AVT.FVT),
  Aachen 52074,
  Germany
}
\def\TUD{
  Delft University of Technology,
  2629 HZ, Delft,
  The Netherlands
}

\newcommand{\mytitle}{Physics-Informed Neural Networks for Dynamic Process Operations with Limited Physical Knowledge and Data}

\newcommand{\affil}{
  \begin{itemize}[leftmargin=3mm, itemsep=0mm]
    \item[$^a$]\IEK10
    \item[$^b$]\RWTH
    \item[$^c$]\JARA
    \item[$^d$]\SVT
    \item[$^e$]\FVT
    \item[$^f$]\TUD
  \end{itemize}
}

\def\firstAuthor{Mehmet Velioglu}

\newcommand{\myauthor}{\firstAuthor$^{a,b}$,
Song Zhai$^{e}$,
Sophia Rupprecht$^{a,f}$,
Alexander Mitsos$^{c,a,d}$,
Andreas Jupke$^{e}$,
Manuel Dahmen$^{a,*}$
}
\newcommand{\correspondingauthor}{M. Dahmen, \IEK10 \newline E-mail: m.dahmen@fz-juelich.de}
\author{\myauthor}

\usepackage[
  colorlinks,
  linkcolor=blue,
  citecolor=blue,
  urlcolor=blue,
  pdftitle={\mytitle},
  pdfauthor={\firstAuthor}
]{hyperref}
\usepackage{svg}
\usepackage[capitalise, nameinlink]{cleveref}
\crefname{table}{Tab.}{Tab.}

\begin{document}

\doublespacing

\thispagestyle{firststyle}

\begin{center}
    \begin{large}
      \textbf{\mytitle}
    \end{large} \\
    \myauthor
\end{center}

\vspace{0.5cm}

\begin{footnotesize}
    \affil
\end{footnotesize}

\vspace{0.5cm}
    
\begin{abstract}
   In chemical engineering, process data are expensive to acquire, and complex phenomena are difficult to fully model. We explore the use of physics-informed neural networks (PINNs) for modeling dynamic processes with incomplete mechanistic semi-explicit differential-algebraic equation systems and scarce process data. In particular, we focus on estimating states for which neither direct observational data nor constitutive equations are available. We propose an easy-to-apply heuristic to assess whether estimation of such states may be possible. As numerical examples, we consider a continuously stirred tank reactor and a liquid-liquid separator. We find that PINNs can infer immeasurable states with reasonable accuracy, even if respective constitutive equations are unknown. We thus show that PINNs are capable of modeling processes when relatively few experimental data and only partially known mechanistic descriptions are available, and conclude that they constitute a promising avenue that warrants further investigation. 
\end{abstract}

\vspace{0.5cm}


\vspace{0.5cm}

\vspace*{5mm}

\newpage

\section{Introduction}\label{sec:intro} 
Dynamic operation and control of chemical and biotechnological processes are essential for efficient and sustainable production. Mathematical models describing the behavior of such processes are often classified concerning their degree of reliance on physical/chemical knowledge or data into three categories: (1) white-box or first-principle or mechanistic models, (2) black-box or data-driven models, and (3) gray-box or hybrid models \citep{Zendehboudi2018ApplicationsReview, MARQUARDT1996591}.\par 

Black-box modeling relies on (measurement) data to establish a predictive relation between process inputs and outputs, thus avoiding the need for a mechanistic process description. In recent years, approaches involving deep neural networks (DNNs) have become particularly prominent data-driven models for process operations. DNNs can model nonlinear dependencies between multiple inputs and outputs \citep{Goodfellow2016DeepLearning} but require extensive training data and often fail to make physically consistent predictions in scientific or engineering applications \citep{Zendehboudi2018ApplicationsReview}. 
In contrast, mechanistic process models are based on the governing physical and chemical laws of a system and suitable constitutive equations and comprise relatively few parameters that need to be estimated from data \citep{vonStosch2014HybridFuture}. They typically allow for physically consistent predictions. However, in chemical and biotechnological processes, complex phenomena such as reaction kinetics, coalescence, or sedimentation often lack a rigorous mathematical description, hindering the mechanistic modeling of such processes \citep{Kahrs2008IncrementalModels}. Hybrid modeling combines mechanistic and data-driven modeling and aims to take advantage of the respective strengths and mitigate the respective weaknesses of the two approaches. Compared to purely data-driven models, suitably-designed hybrid models require less training data, make physically more consistent predictions, and (thus) extrapolate to a higher extent \citep{Kahrs2007TheOptimization}. 

Hybrid models have been used extensively to model dynamic process operation problems if complete system knowledge is unavailable \citep{Roffel2006ProcessPrediction} and thus have become a crucial modeling tool for numerous tasks related to chemical process control \citep{Asprion2019Gray-BoxProcesses}. Various types of hybrid model structures have been proposed over the years in the process systems engineering (PSE) community, with the sequential approach and the parallel approach being the most prominent structures. For instance, \citet{Psichogios1992AModeling} studied incorporating an artificial neural network to predict states lacking a constitutive description inside an otherwise mechanistic model for a fed-batch bioreactor (sequential approach). \citet{Su1992IntegratingModeling} proposed to correct the mismatch between a white-box model and process data from a polymer reaction system by a neural network (parallel approach). The parallel approach can also be combined with the sequential approach, i.e., a second mechanistic model is added after the parallel hybrid model to enforce physically consistent predictions, see, e.g., \citep{Thompson1994ModelingNetworks}. Recently, the popularity of hybrid modeling in chemical engineering has been increasing again due to advancements in machine learning and the rise of digital twins in smart manufacturing \citep{YANG2020106874}. Some notable contemporary works on hybrid modeling are dedicated to the estimation of (spatio-)temporally varying parameters, which is related to the estimation of states with missing constitutive equations, the main topic of our article. Specifically, \citet{SHAH2022135643} estimate time-varying parameters in fermentation processes, \citet{PahariLatent} estimate spatio-temporally varying diffusivity in a reaction-diffusion model, and \citet{sitapurehybrid} estimate kinetic parameters in a batch crystallization process with a transformer architecture.  
For further applications of hybrid modeling in chemical engineering, we refer the reader to review papers by  \citet{SANSANA2021107365, YANG2020106874, Sharma_2022, SchweidtmannMLCE}.

Physics-based regularization of DNNs gives rise to so-called physics-informed neural networks (PINNs), which have some similarities to hybrid models but are better regarded as a special variant of a data-driven model that is trained with available physical laws as constraints \citep{BRADLEY2022107898}. Specifically, in a PINN, the DNN acts as the sole prediction model, but it is informed about governing physical laws during training through additional terms in the loss function \citep{Nabian2019Physics-DrivenAnalysis, 2021Nat}. In contrast, hybrid models have distinct mechanistic and data-driven sub-models which jointly produce a prediction \citep{BRADLEY2022107898, SCHWEIDTMANN2024100136}. 

The origins of physics-based regularization date back to (at least) the works of \citet{Lagaris1998ArtificialEquations} on solving ordinary and partial differential equations using neural networks (NNs) as universal function approximators. This approach was originally not taken up widely, likely due to the general limitations of NN training at that time. However, \citet{Raissi2019Physics-informedEquations} recently revisited the physics-based regularization approach using modern algorithms and tools for training and introduced the term PINN. 

The original PINN architectures \citep{Raissi2019Physics-informedEquations, Nascimento2020ANetwork} did not account for varying initial/boundary conditions or control inputs. However, \citet{Antonelo2021Physics-InformedSystems} showed that adding control inputs and initial conditions to the NN makes the PINN approach suitable for control applications. 
Another application of PINNs for control purposes was proposed by \citet{Arnold2021StatespaceNetworks}, who pursued a state-space modeling approach based on PINNs, including initial conditions as inputs to the NN. However, separate networks are trained for each discretized control actuation instead of adding control inputs to the network.

Recently, PINNs have also seen a surge in chemical engineering applications, mainly in the form of physics-informed \emph{recurrent} neural networks \citep{ZHENG2023103005}. For instance, they have been applied in conjunction with model predictive control (MPC) to a continuously stirred tank reactor (CSTR) \citep{ZHENG2023103005} and a batch crystallization process \citep{WU2023556}, to control systems with noisy data \citep{ALHAJERI202234} and parametric uncertainty \citep{ZhengPINN}, and to fluid flow problems, most notably flow field prediction in cyclone separators \citep{QUEIROZ2021100002} and a Van de Vusse CSTR \citep{choi2022physics}. \citet{JiStiffPINN} developed PINNs that can address stiff chemical kinetic problems. 

While studies have shown that PINNs are promising model candidates for chemical engineering applications, open questions remain about their utility for state estimation. In general, state estimation is concerned with estimating the state of a given process utilizing measurement data and a mathematical process model \citep{Barfoot_2017, gelb1974applied}. 
State estimation is often performed with filtering techniques, e.g., the Kalman filter \citep{Kalman1960AProblems}, which have recently also been combined with PINNs, see, e.g., \citep{tan2023vehicle, Arnold2021StatespaceNetworks}. 
PINNs have also been used to estimate unmeasured states directly, i.e., without the use of a state estimation technique. For instance, \citet{Raissi2020HiddenVisualizations} estimated velocity and pressure fields from the concentration data of a passive scalar from flow field visualizations, using Navier Stokes equations as the physics knowledge. Recently, \citep{WU2023556} showed that PINNs with partial physics knowledge can estimate immeasurable states in a batch crystallization process by using the known governing equations of these states. The question, however, remains whether PINNs can estimate states for which neither direct observational data nor constitutive equations are available.

In the present work, we thus set out to answer the following two questions: (i) Can PINNs estimate immeasurable process states for which constitutive equations are not known? (ii) Under which conditions can we expect this to work? To this end, we will first conceptualize PINN-based dynamic process models in a setting of partially known mechanistic equations as well as measured and unmeasured process states. Specifically, we consider systems that (i) can be described by differential-algebraic equations (DAEs) in principle, (ii) for which only partial mechanistic knowledge in the form of some known equations is available, and (iii) for which process data for some states is available.
Regarding the PINN modeling, we follow the standard approach, as it was first introduced by \citet{Raissi2019Physics-informedEquations}, but with the extensions to initial states and control inputs by \citet{Antonelo2021Physics-InformedSystems}. 
We propose the use of an incidence matrix as an easy-to-apply heuristic to \emph{a priori} evaluate whether estimation of unmeasured states with a PINN may be possible. 
We then perform extensive numerical studies by using two fully-known mechanistic models to emulate situations where some, but not full, mechanistic knowledge is available for modeling purposes.
Specifically, we study a CSTR model with Van de Vusse reaction from the literature \citep{vandeVusse1964Plug-flowReactor} and a liquid-liquid separator for which we develop a model by extending the model from \cite{Backi2018AFirst-Principles,Backi.2019}. 
We follow an in-silico approach to generate process data, i.e., we use the full-order mechanistic model, which in a real situation would not be available, to generate synthetic observational data. Controlling the amount and diversity of training data allows us to run extensive numerical experiments on the fitting and generalization capabilities of PINNs as well as vanilla neural network benchmark models, i.e., multilayer perceptrons. Following the taxonomy of process quantities and model equations by \citet{MARQUARDT1996591}, we distinguish balance equations and constitutive equations and emulate situations with different degrees of mechanistic knowledge available for PINN model development.

The paper is structured as follows: Section \ref{sec:method} presents the proposed approach for PINN-based dynamic process modeling with incomplete physical knowledge, and our heuristic for assessing the state estimation capabilities of a PINN. Section \ref{sec:results} provides numerical examples and results for the CSTR, focusing on the physics-informed part of the PINN by varying the amount of physical knowledge provided. Section \ref{sec:Settler} provides numerical examples and results for the liquid-liquid separator, focusing on the data-driven part of the PINN by varying the number of measured properties provided as NN inputs. In all examples, the empirical findings are related to the results from the heuristic. Section \ref{sec:conclusion} discusses the conclusion and future work.

\section{Methods}\label{sec:method}
\subsection{Preliminaries}
\citet{Raissi2019Physics-informedEquations} introduced PINNs to find data-driven solutions to partial differential equations (PDEs) utilizing DNNs. In their approach, they employ the NN to approximate the solution of a PDE problem. The inputs to the DNN are the spatio-temporal coordinates, and the DNN outputs are the states of the dynamic system. The DNN is trained in a semi-supervised manner, e.g., with small amounts of labeled data, i.e., process data with corresponding input/output relations, and large amounts of unlabeled data, i.e., collocation points in time and space where residuals of governing equations, i.e., the PDEs, are computed. These residuals constitute a loss term that penalizes the deviations of the DNN outputs from the governing equations. Thus, PINNs can learn to obey the physical laws of the system. 

\begin{figure}
    \centering
    \includegraphics[scale=0.5]{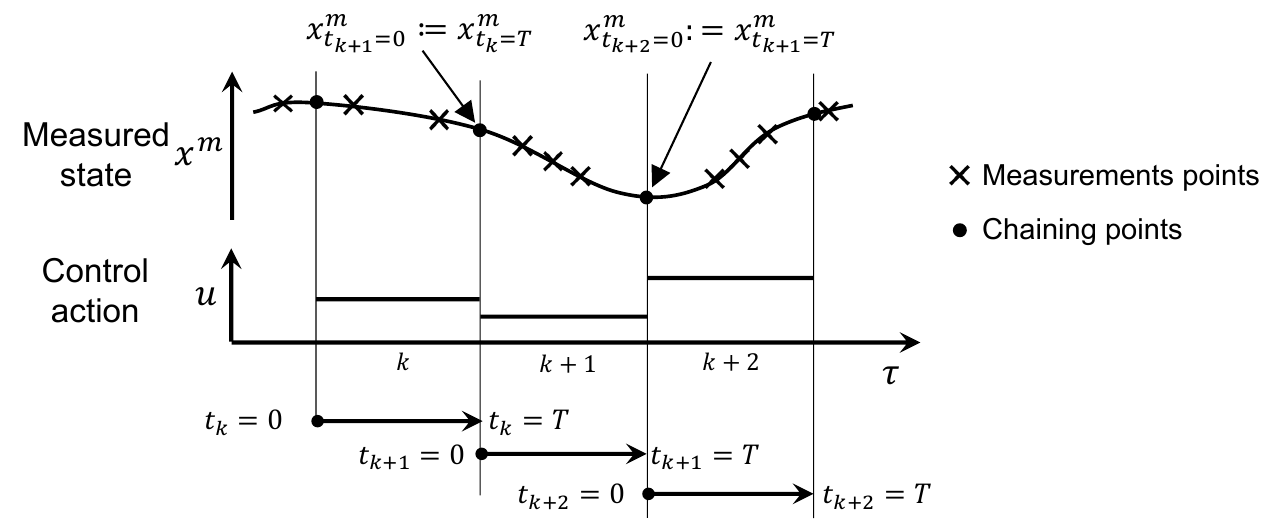}
    \caption{Relationship between PINN time $t$ and process time $\tau$: The PINN time domain $[0,T]$ corresponds to the length of a step-wise constant control input. In general, PINN time $t$ differs from process time $\tau$ and chaining of model predictions is required to simulate longer periods of time. Only if the control input is constant over the entire process duration, $t$ and $\tau$ coincide. Measurements can come from an irregular grid.}
    \label{fig:step-wise-control}
\end{figure}

In their original form, PINNs do not account for control variables. The extension to control applications is, however, straightforward: \citet{Antonelo2021Physics-InformedSystems} added the control variable(s) and initial states as NN inputs. Considering initial states as network inputs means that the PINN model can be trained for various samples of initial states and control variables, facilitating extensive coverage of the state and control action spaces. 
The time domain of the PINN can be chosen according to the needs of the control scheme,  e.g., in MPC applications, step-wise constant control inputs are often used. Thus, if the PINN time domain $[0,T]$ corresponds to the length of a step-wise constant control input, the control inputs from the perspective of the NN are not functions of time but constants. It is therefore, in general, necessary to distinguish PINN time $t$ from process time $\tau$ and to chain the PINN predictions in order to simulate longer periods involving changing control inputs (cf. Figure \ref{fig:step-wise-control}). Note that in the numerical examples in Sections \ref{sec:results} and \ref{sec:Settler}, for the sake of a simple implementation, we study varying control inputs which are however kept constant throughout the entire process duration, thus implying $t=\tau$. 
For further details on including control actions into PINNs, we refer the reader to \citet{Antonelo2021Physics-InformedSystems} for integrating PINNs into MPC. %

\subsection{PINN-based dynamic process modeling with partial physical knowledge}

We consider the scenario where a partial mechanistic process model is available that can be used for physics-based regularization of a NN. We assume that this partial process model comes in the form of a semi-explicit differential-algebraic equation (DAE) system \citep{brenan1996numerical}:
\begin{subequations}
    \begin{align}
        &\Dot{\Vx}(t) = \Ff(\Vx(t), \Vy(t), \Vu), \label{eq:exp-dae-1}\\ 
        &\mathbf{0} = \Fg(\Vx(t), \Vy(t), \Vu) \label{eq:exp-dae-2} 
    \end{align}
    \label{eq:explicit-dae}
\end{subequations}
Here, $\Vx(t) \in \mathbb{R}^{n_x}$ is the differential states vector, $\Vy(t) \in \mathbb{R}^{n_y}$ is the algebraic states vector, and $\Vu \in \mathbb{R}^{n_u}$ is the control inputs vector.  
The dot symbol ( $\Dot{}$ ) denotes a time derivative. 
$\bm f$ denotes the right-hand side (RHS) of the ordinary differential Equations \ref{eq:exp-dae-1}, and $\bm g$ is the RHS of the algebraic Equations \ref{eq:exp-dae-2}.

In a practical setting, some states might be impossible to measure (immeasurable), e.g., reaction rate constants, or some states might be impractical/expensive to measure, e.g., concentrations. The term \emph{unmeasured states} covers both of these types and will be used throughout this work. We aim to estimate unmeasured process states with the available partial mechanistic knowledge and measurement data on other measured states. To this end, we sub-categorize the differential and algebraic states into measured and unmeasured states, using superscripts $m$ and $u$, respectively. This is a special case of the more general output equations used in observability analysis and control, see, e.g., \cite{lee1967foundations}.

To predict the measured states $\Vx^m(t) \in \mathbb{R}^{n_{x^m}}$, $\Vy^m(t) \in \mathbb{R}^{n_{y^m}}$ and to estimate the unmeasured states $\Vx^u(t)  \in \mathbb{R}^{n_{x^u}}$, $\Vy^u(t)  \in \mathbb{R}^{n_{y^u}}$, we use the neural network $\textbf{NN}_{\textbf{w},\textbf{b}}$ with weights $\textbf{w}$ and biases $\textbf{b}$, i.e., $[\Vxt(t),\Vyt(t)] = \textbf{NN}_{\textbf{w},\textbf{b}}(t, \Vx^m(t_0), \Vu)$, where $\Vxt(t)$ and $\Vyt(t)$ denote the NN predictions of the differential and algebraic states, respectively. The network inputs are the time $t$, the initial values of the measured differential states $\Vx^m(t_0)$, and the control inputs $\Vu$.
The NN parameters $\textbf{w}$ and $\textbf{b}$ can be learned by minimizing the mean squared error loss, similar to \cite{Raissi2019Physics-informedEquations, Antonelo2021Physics-InformedSystems}:
\begin{subequations}
    \label{eq:pinn-loss-ic-cv}
    \begin{align}
    MSE_{total} &=  MSE_{data} + \lambda_1 MSE_{physics} + \lambda_2 MSE_{init}, \label{eq:totalloss} \\
    \label{eq:data-loss}MSE_{data} &= \frac{1}{n_{x^m}N_d} \sum_{j=1}^{N_d} (\Vxt^m(t_j) -  \Vx^m(t_j))^2 + \frac{1}{n_{y^m}N_d} \sum_{j=1}^{N_d} (\Vyt^m(t_j) - \Vy^m(t_j))^2, \\
    \begin{split}
    \label{eq:res-loss}MSE_{physics}  &= \frac{1}{n_x N_e} \sum_{j=1}^{N_e} \left(\Dot{\Vxt}(t_j) - {\bm f}(\Vxt(t_j), \Vyt(t_j), \Vu_j)\right)^2 + \\ & \quad \frac{\lambda_g}{n_y N_e} \sum_{j=1}^{N_e} \left({\bm g}(\Vxt(t_j), \Vyt(t_j), \Vu_j)\right)^2,         
    \end{split} \\
    \label{eq:init-loss}MSE_{init} &= \frac{1}{n_{x^m}N_i} \sum_{j=1}^{N_i} (\Vxt^m_j(t_0) - \Vx^m_j(t_0))^2
    \end{align}
\end{subequations}

Here, $MSE_{data}$ corresponds to the loss term accounting for the measurement data, $MSE_{physics}$ corresponds to the loss term that is computed with the available physics knowledge (Equations \ref{eq:exp-dae-1} and \ref{eq:exp-dae-2}), and $MSE_{init}$ corresponds to a loss term that describes the mismatch between the NN predictions at $t = t_0$ and the initial values ${\Vx}^m_j(t_0)$. $N$ denotes the number of data points. Note that the subscript $j$ refers to finitely many samples taken at times $t_j$, with corresponding initial values ${\Vx}_j^m(t_0)$ and control actions $\Vu_j$. We omit the latter two from the notation for simplicity. The subscripts $d$, $e$, and $i$ correspond to data points associated with $MSE_{data}$, $MSE_{physics}$, and $MSE_{init}$, respectively. 

$\lambda_1$ and $\lambda_2$ denote the weights of the physics and initial condition loss terms, respectively, and $\lambda_g$ establishes a weighting between the algebraic and the differential equations in the physics loss term. 

Note that for the calculation of $MSE_{physics}$ and $MSE_{init}$ no measurement data are needed. For $MSE_{physics}$, we calculate the physics residuals using Equations \ref{eq:exp-dae-1} and \ref{eq:exp-dae-2} at randomly sampled time points $t = t_j$. For $MSE_{init}$, we train the NN predictions $\Vxt^m(t=t_0)$ to comply with the initial values ${\Vx}^m(t_0)$, again for randomly sampled values in a given range.

\begin{figure}[h]
\centering
    \centering
    \includegraphics[scale=0.5]{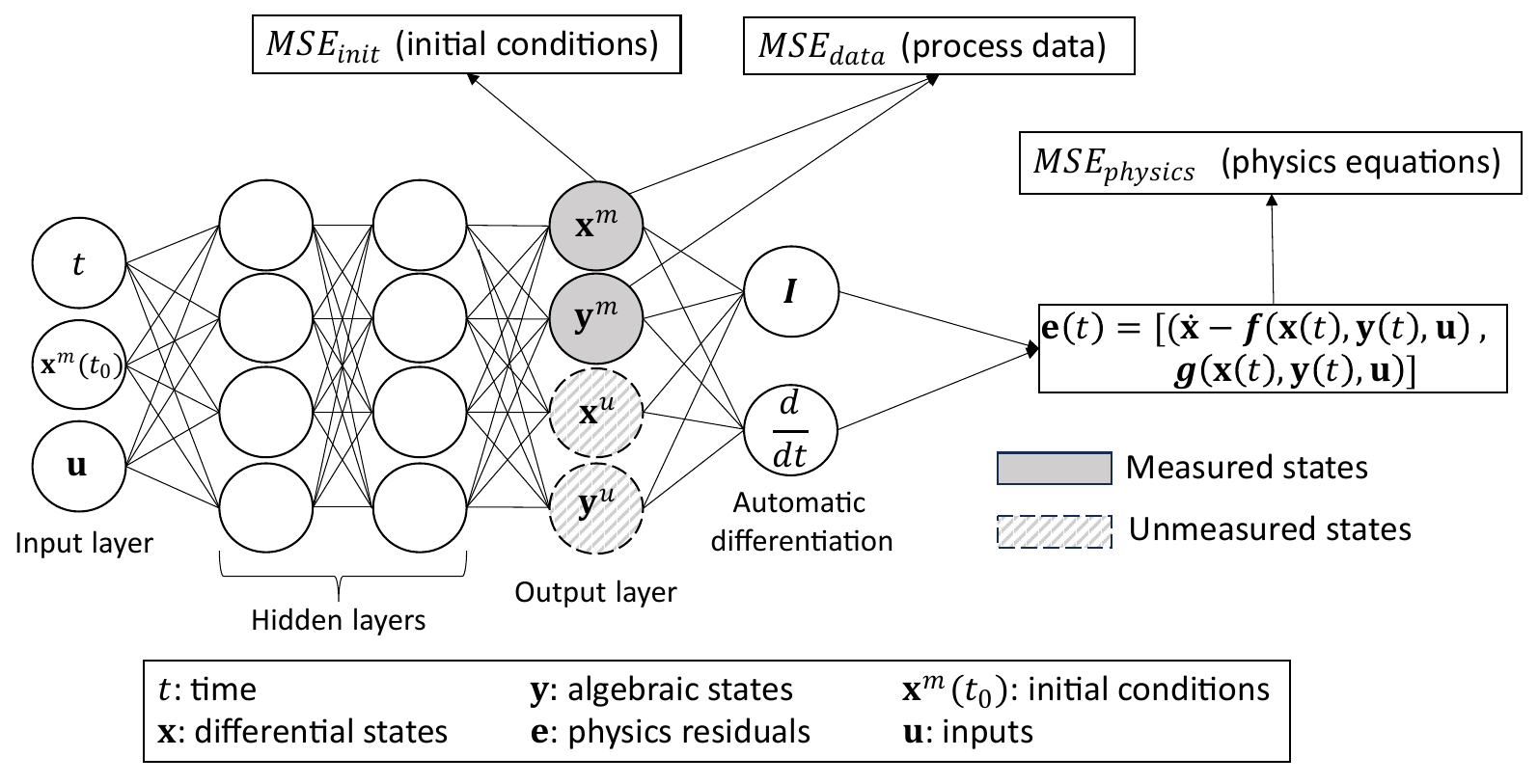}
    \caption{PINN-based dynamic process model with semi-explicit DAE physics model}
    \label{fig:general-pinn-model}
\end{figure}

\subsection{Heuristic for assessing PINN state estimation capabilities}
\label{method-conceptual}

We propose a heuristic to a priori assess whether a PINN may be capable of estimating unmeasured process states by drawing inspirations from DAE solvability analysis, see, e.g.,  \citep{brenan1996numerical}. Our conjecture is that the PINN can leverage training data, i.e., samples for ${\Vx}_m(t_j)$ and ${\Vy}_m(t_j)$, to ``solve'' the known Equations \eqref{eq:exp-dae-1} and \eqref{eq:exp-dae-2} for the unknown states ${{\Vx}}_u(t_j)$ and ${{\Vy}}_u(t_j)$ at a point $t_j$. 
Specifically, our heuristic mimics structural index analysis by means of an incidence matrix \citep{duff1986computing,Gani1992,Unger1995StructuralApplications}. 
In our PINN incidence matrix, the rows represent the RHSs of the known physics equations, i.e., $\Ff$ and $\Fg$ (see Equations \eqref{eq:exp-dae-1} and \eqref{eq:exp-dae-2}), and the columns represent the unmeasured process states ${\Vx}_u$ and ${\Vy}_u$. Each occurrence of an unmeasured state in $\bm f$ and $\bm g$ is indicated by drawing a cross ($\cross$) in the corresponding entry of the matrix. 
Note that the PINN uses AD to compute $\dot{\Vxt}$, i.e., the derivative of the NN outputs $\Vxt(t)$ with respect to the NN input $t$. Moreover, the NN learns to assemble state trajectories from the data provided at distinct time points $t_j$, and thus, it implicitly learns time-derivatives of the states. Consequently, we do not consider $\dot{\Vx}$, i.e., the left-hand side (LHS) of Equations \ref{eq:exp-dae-1}, as unknowns but restrict our analysis to the RHSs $\bm f$ and $\bm g$ where no time-derivatives appear (see Equations \eqref{eq:exp-dae-1} and \eqref{eq:exp-dae-2}). 
This implies that we do not consider $\dot{x}_j$ as an occurrence of $x_j$ when we assemble the incidence matrix. 

We conjecture that the incidence matrix having a full-column rank, i.e., if exactly one cross in each column can be marked with a circle without marking more than one cross in a single row, constitutes an indicator for possible state estimation. A simple example of an incidence matrix for a PINN is given in Table \ref{tab:incidence}. Note that an incidence matrix having more equations than unmeasured states, i.e., more rows than columns, is not a concern in itself. In fact, each additional equation may provide additional regularization to the NN and thus may be regarded as beneficial. 
We stress that the incidence matrix is a heuristic, i.e., it represents neither a necessary nor a sufficient condition for state estimation with a PINN (see Sections SM5 and SM6 of the Supplementary Materials), and thus, it can give wrong results. 
Note that for fully-specified dynamic systems, necessary and sufficient criteria for observability analysis exist, see, e.g., \citep{lee1967foundations,kou1973observability}, based on trajectory information.
Since we have an incomplete physics model, we instead construct the heuristic with a point-wise analysis, similar to the solvability analysis of equation systems \citep{brenan1996numerical,duff1986computing,Gani1992,Unger1995StructuralApplications}.
The practical construction and interpretation of the incidence matrix are demonstrated extensively in Sections \ref{sec:results} and \ref{sec:Settler}. 

\begin{table}[h]
\centering
\caption{Incidence matrix for a PINN with a semi-explicit DAE physics model: Measurement data for training is available for $x_1^m$ only. The unmeasured states $x_2^u$ and $y^u$ shall be estimated from the data on $x_1^m$. The cross ($\cross$) denotes the occurrence of an unmeasured state in a physics equation. The incidence matrix has full-column rank, as it is possible to mark exactly one cross in each column without marking more than one cross in a single row. }
\label{tab:incidence}
\begin{tabular}{ cc } 
Known physics model (semi-explicit DAE) & Incidence matrix \\  
\begin{tabular}{l l l}
(a): & $\dot{x}_1^m$ & $= x_1^m + x_2^u$ \\
(b): & $\dot{x}_2^u$ & $= 3x_1^m$ \\
(c): & $0$ & $=x_1^m x_2^u + y^u$ \\ 
\end{tabular} &  
\begin{tabular}{ c | c | c } 
$[\bm f, \bm g]\downarrow \quad [\Vx^u, \Vy^u]\rightarrow$ & $x_2^u$ & $y^u$ \\
\hline
(a) & $\otimes$ \\
\hline
(b) & \\
\hline
(c) & $\cross$ & $\otimes$ \\
\end{tabular} \\
\end{tabular}
\end{table}

\subsection{Vanilla NN benchmark models}
To compare the predictions of a PINN model with a purely data-driven benchmark, we choose a feed-forward artificial neural network (ANN), as ANNs are widely used and can have a similar network architecture as the PINN model, thus allowing us to study the effects of the physics-based regularization. To make the comparison as meaningful as possible, we use the same hyperparameters and training scheme for the PINN model and the vanilla ANN model. Still, the network architecture for the vanilla ANN is slightly different from that of the PINN in the sense that only the measured states can be network outputs, as no process data is available for the unmeasured states. We use the following loss function to train the vanilla ANN, omitting the physics-based regularization term in Equation \eqref{eq:totalloss} but keeping the loss term for the initial conditions:
\begin{align*}
    \label{eq:mlp-loss}
    \begin{split}
    MSE &=  MSE_{data} + \lambda_1 MSE_{init}, \\
    MSE_{data} &= \frac{1}{n_{x^m}N_d} \sum_{j=1}^{N_d} (\Vxt^m(t_j) -  \Vx^m(t_j))^2 + \frac{1}{n_{y^m}N_d} \sum_{j=1}^{N_d} (\Vyt^m(t_j) - \Vy^m(t_j))^2, \\
    MSE_{init} &= \frac{1}{n_{x^m}N_i} \sum_{j=1}^{N_i} (\Vxt^m_j(t_0) - \Vx^m_j(t_0))^2
    \end{split}
\end{align*}
  
\section{Numerical example 1: Van de Vusse Reactor}\label{sec:results}

We use the Van de Vusse \citep{vandeVusse1964Plug-flowReactor} CSTR, a common benchmark problem in the literature on nonlinear control applications \citep{Chen1995NonlinearCSTR}, to investigate generalization, state estimation, and extrapolation capabilities of the PINN models under varying amounts of physical knowledge provided through physics equations. Thus, we focus on the physics regularization aspect of the PINN. 

\begin{figure}
    \centering
    \includegraphics[scale=0.5]{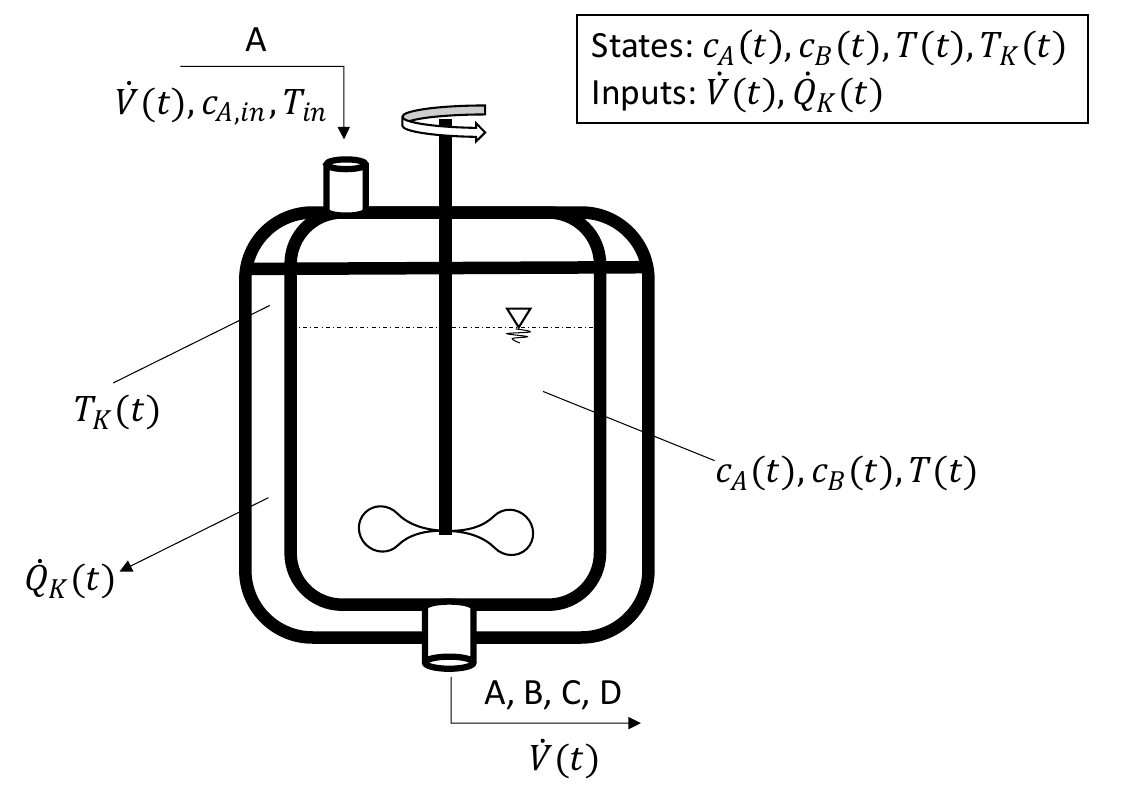}
    \caption{Schematic representation of the van de Vusse CSTR}
    \label{fig:vdv-reactor-image}
\end{figure}

The van de Vusse reaction scheme reads:
\begin{align*}
    \begin{split}
    &\ce{A ->[k1] B ->[k2] C}, \\
    &\ce{2A ->[k3] D}.
    \end{split}
\end{align*}
Substance $A$ is fed to the reactor with concentration $c_{A,in}$ and temperature $T_{in}$. Substance $B$ is the desired product, whereas substances $C$ and $D$ are unwanted byproducts. Heat is removed from the cooling jacket fluid with rate $\Dot{Q}_K$ by an external heat exchanger. The schematic of the CSTR is given in Figure \ref{fig:vdv-reactor-image}.
The dynamics of the reactor are given by the following nonlinear equations derived from component balances for substances $A$ and $B$ and energy balances for the reactor and the cooling jacket \citep{Chen1995NonlinearCSTR}:
\begin{subequations}
    \begin{align}
        \Dot{c}_A(t) &= \frac{\Dot{V}(t)}{V_R} (c_{A,in} - c_A(t)) - k_1(T) c_A(t) - k_3(T) c_A(t)^2, \\
        \Dot{c}_B(t) &= -\frac{\Dot{V}(t)}{V_R} c_B(t) + k_1(T) c_A(t) - k_2(T) c_B(t), \\
        \begin{split}
        \Dot{T}(t)   &= \frac{\Dot{V}(t)}{V_R} ( T_{in} - T(t)) - \frac{1}{\rho C_p} [k_1(T) c_A(t) \Delta H_{AB} + k_2(T) c_B(t) \Delta H_{BC} \\
                     &\qquad + k_3(T) c_A(t)^2 \Delta H_{AD}] + \frac{k_w A_R}{\rho C_p V_R} (T_K(t) - T(t)), \label{eq:vdv-T}
        \end{split}
        \\
        \Dot{T}_K(t) &= \frac{1}{m_K C_{pK}}[\Dot{Q}_K(t) + k_w A_R (T(t) - T_K(t))] \label{eq:vdv-TK}
    \end{align}
\label{eq:vdv}
\end{subequations}
Here, $c_A(t)$ and $c_B(t)$ denote the concentrations of substances A and B, $T(t)$ is the reactor temperature, and $T_K(t)$ is the cooling jacket temperature, assumed to be uniform in space. The aforementioned quantities correspond to the differential states $\Vx$ of the Van de Vusse CSTR, i.e., $\Vx = [c_A, c_B, T, T_K]^T$. The flow rate $\Dot{V}(t)$, and the heat transfer rate by the coolant $\Dot{Q}_K(t)$ (heat removal) are the manipulated variables. Note that the dot notation in $\Dot{V}(t)$ and $\Dot{Q}_K(t)$ indicates flow rates (as opposed to time derivatives). The reaction rate constants $k_i(T)$ correspond to the algebraic states $\Vy$ and are calculated using the Arrhenius equation:
\begin{equation}
    k_i(T) = k_{i0} \exp (\frac{E_{a,i}}{T}), \qquad i = 1,2,3
\label{eq:arrhenius}
\end{equation}
All parameters listed in Equations \eqref{eq:vdv} and \eqref{eq:arrhenius} are given in Table \ref{tab:vdv-param}.

\begin{table}[h]
\centering
\ra{1.3}
\caption{Parameters for the van de Vusse CSTR, taken from \citep{Chen1995NonlinearCSTR}.}
\begin{tabular}{lll}
\toprule 
Parameter & Symbol & Value\\
\midrule
$\textrm{inlet molar flow rate of substance A}$ & $c_{A,\textrm{in}}$ & \SI{5.10}{\mol\per\L} \\
$\textrm{inlet temperature}$ & $T_{\textrm{in}}$ & \SI{378.1}{\kelvin} \\
$\textrm{collision factor for reaction 1}$ & $k_{10}$ & \SI{1.287e12}{1\per\hour} \\
$\textrm{collision factor for reaction 2}$ & $k_{20}$ & \SI{1.287e12}{1\per\hour} \\
$\textrm{collision factor for reaction 3}$ & $k_{30}$ & \SI{9.043e9}{\litre\per\mol\per\hour} \\
$\textrm{activation energy for reaction 1}$ & $E_{a,1}$ & \SI{-9758.3}{\kelvin} \\
$\textrm{activation energy for reaction 2}$ & $E_{a,2}$ & \SI{-9758.3}{\kelvin} \\
$\textrm{activation energy for reaction 3}$ & $E_{a,3}$ & \SI{-8560}{\kelvin} \\
$\textrm{enthalpy of reaction 1}$ & $\Delta H_{AB}$ & \SI{4.2}{\kilo\joule\per\mol_A} \\
$\textrm{enthalpy of reaction 2}$ & $\Delta H_{BC}$ & \SI{-11.0}{\kilo\joule\per\mol_B} \\
$\textrm{enthalpy of reaction 3}$ & $\Delta H_{AD}$ & \SI{-41.85}{\kilo\joule\per\mol_A} \\
$\textrm{density}$ & $\rho$ & \SI{0.9342}{\kg\per\L} \\
$\textrm{heat capacity}$ & $C_P$ & \SI{3.01}{\kilo\joule\per\kg\per\kelvin} \\
$\textrm{heat capacity of coolant}$ & $C_{PK}$ & \SI{2.00}{\kilo\joule\per\kg\per\kelvin} \\
$\textrm{heat transfer coefficient of cooling jacket}$ & $k_w$ & \SI{4032}{\kilo\joule\per\hour\per\meter\squared\per\kelvin} \\
$\textrm{surface area of cooling jacket}$ & $A_R$ & \SI{0.215}{\meter\squared} \\
$\textrm{reactor volume}$ & $V_R$ & \SI{0.01}{\meter\cubed} \\
$\textrm{coolant mass}$ & $m_K$ & \SI{5.0}{\kg} \\
\bottomrule
\end{tabular}
\label{tab:vdv-param}
\end{table}

During our preliminary tests, we observed that having values in a similar order of magnitude for the different PINN inputs and outputs improves the training stability and performance. However, when normalizing the outputs, the PINN physics equations must be scaled accordingly. Thus, we decided to make the time, states, and manipulated variables dimensionless and use dimensionless equations to calculate the physics loss. We give the dimensionless variables and equations in the Supplementary Materials. In addition, we normalize the PINN inputs, i.e., we scale the input features to values between -1 and 1. 

To investigate the effects of varying physical knowledge, we create three different PINN models with increasing physics knowledge. Moreover, we investigate the performance of a vanilla ANN model to facilitate a comparison between the PINN model and a purely data-driven model. We list these models, the physics equations, the knowledge supplied to the PINN model, and the measured and unmeasured states in Table \ref{tab:vdv-models}. Moreover, we give the network schematic of each model in the Supplementary Materials.

\begin{landscape}
        \begin{table}[htbp]
        \caption{Physical knowledge and output configuration (measured and unmeasured process states) for the Van de Vusse CSTR PINN models. In PINN-C, $T$ and $T_K$ can also be unmeasured depending on the case study (cf. Section \ref{sec:diff-se}). The time dependence of the states is not shown explicitly for brevity. The manipulated variables $\frac{\Dot{V}}{V_R}$ and $\Dot{Q}_K$ are the step-wise constant controls which we, for the sake of a simple implementation, keep constant throughout the investigated process duration.}
        \label{tab:vdv-models}
        \begin{tabular}{ c|c|c|c|c }
        \toprule
        \textbf{Model name} & \textbf{Physics knowledge} & \textbf{Physics equations} &  \begin{tabular}{@{}c@{}} \textbf{Measured} \\ \textbf{process states} \end{tabular} & \begin{tabular}{@{}c@{}} \textbf{Unmeasured} \\ \textbf{process states} \end{tabular} \\ \hline
        Vanilla ANN        &              None          &         None    & $c_A$, $c_B$, $T$, $T_K$ & None     \\ \hline
        PINN-A   &        \begin{tabular}{@{}c@{}} Mole balances \\ with net reaction rates \end{tabular}   &   \multicolumn{1}{l|}{                  $\begin{aligned}
                \Dot{c}_A &= \frac{\Dot{V}}{V_R} (c_{A,in} - c_A) + r_A \\
                \Dot{c}_B &= -\frac{\Dot{V}}{V_R} c_B + r_B
            \end{aligned}$    }   & $c_A$, $c_B$, $T$, $T_K$ & $r_A$, $r_B$          \\ \hline
        PINN-B      &     \begin{tabular}{@{}c@{}} Mole and energy balances \\  with individual reaction rates \end{tabular}        &               
        \multicolumn{1}{l|}{$\begin{aligned}
                \Dot{c}_A &= \frac{\Dot{V}}{V_R}(c_{A,in} - c_A) - r_1  - r_3 \\
                \Dot{c}_B &= -\frac{\Dot{V}}{V_R} c_B + r_1 - r_2 \\
                \Dot{T}   &= \frac{\Dot{V}}{V_R} ( T_{in} - T) + \frac{k_w A_R}{\rho C_p V_R} (T_K - T) \\
                &- \frac{1}{\rho C_p} (r_1 \Delta H_{AB} + r_2 \Delta H_{BC} + r_3 \Delta H_{AD}) \\
                \Dot{T}_K &= \frac{1}{m_K C_{pK}}(\Dot{Q}_K + k_w A_R (T - T_K))
            \end{aligned}$  }   & $c_A$, $c_B$, $T$, $T_K$ & $r_1$, $r_2$, $r_3$                   \\ \hline
        PINN-C   &  \begin{tabular}{@{}c@{}} Mole and energy balances \\ with reaction rate expressions \\ (without Arrhenius' law) \end{tabular}     &         
                \multicolumn{1}{l|}{$\begin{aligned}
                \Dot{c}_A &= \frac{\Dot{V}}{V_R} (c_{A,in} - c_A) - k_1 c_A - k_3 c_A^2\\
                \Dot{c}_B &= -\frac{\Dot{V}}{V_R} c_B + k_1 c_A - k_2 c_B\\
                \Dot{T}   &= \frac{\Dot{V}}{V_R} ( T_{in} - T) + \frac{k_w A_R}{\rho C_p V_R} (T_K - T) \\
                &-  \frac{1}{\rho C_p} (k_1 c_A \Delta H_{AB} + k_2 c_B \Delta H_{BC} + k_3 c_A^2 \Delta H_{AD}) \\
                \Dot{T}_K &= \frac{1}{m_K C_{pK}}(\Dot{Q}_K + k_w A_R (T - T_K)) 
                \end{aligned} $}  & $c_A$, $c_B$, $T$, $T_K$  & $k_1$, $k_2$, $k_3$                           \\ \bottomrule
        \end{tabular}
        \end{table}
\end{landscape}

\subsection{Data set generation, training, and hyperparameter selection}
\label{sec:vdv-data}
We assume the operating ranges presented in Table \ref{tab:vdv-ranges}, with a selected time interval for step-wise control changes of $T=60 \thinspace \textrm{s}$, i.e., $t \in [0, 60] \thinspace \textrm{s}$. Data generation to calculate the physics loss term $MSE_{physics}$ and the initial condition loss term $MSE_{init}$ in Equations \eqref{eq:pinn-loss-ic-cv} are done by selecting $N_e = 10,000$ collocation and $N_i=100$ initial value points. This selection is done using Latin Hypercube sampling \citep{Iman1981AnAssessment}.  

For the process data generation, we use the explicit Runge-Kutta method of order 5, utilizing \textit{solve\_ivp} solver from \textit{scipy.integrate} module in Python \citep{Virtanen2020SciPyPython,Dormand1980AFormulae}. We solve the full-order process model (Equations \eqref{eq:vdv} and \eqref{eq:arrhenius}) for time $t \in [0, 60]\thinspace \textrm{s}$ with random inputs for $c_A(t_0)$, $c_B(t_0)$, $T(t_0)$, $T_K(t_0)$, $\frac{\Dot{V}}{V_R}$, $\Dot{Q}_K$ in the given ranges and keeping the manipulated variables $\frac{\Dot{V}}{V_R}$ and $\Dot{Q}_K$ constant throughout the investigated process duration of $60\thinspace \textrm{s}$, with relative and absolute error of \num{1e-13} and \num{1e-16} respectively. We output each process trajectory on an equidistant time grid with step-size $\Delta t = 0.6 \thinspace \textrm{s}$. Note that $\Delta t = 0.6 \thinspace \textrm{s}$ pertains to the granularity of the training/testing trajectories; the PINN at the prediction phase can make one-shot predictions for any time $t \in [0, 60] \thinspace \textrm{s}$. Moreover, the training/testing trajectories could also be obtained from an irregular, i.e., non-equidistant, time grid. We create $N_{total} = 100$ trajectories from which we select $N_{test} = 20$ trajectories for testing. For training, we use $N_{train}$ trajectories, each one having $N_m = 101$ data points. The total number of measurement points is thus $N_d = N_{train} N_m$. Specifically, we create two training sets from the 80 trajectories that are not used for the testing: First, we create a training set representing a \textit{low-data regime} consisting of only $N_{train} = 20$ training trajectories. Second, we create a training set representing a \textit{high-data regime} consisting of $N_{train} = 80$ training trajectories.

\begin{table}[h]
\centering
\caption{Operating ranges for states and inputs in the Van de Vusse CSTR example. The lower bound is denoted by $\textrm{lb}$, and the upper bound is denoted by $\textrm{ub}$. These values are chosen to remain in the vicinity of a steady state. Extreme values refer to the minimum and maximum values appearing in a generated trajectory.}
\begin{tabular}{@{}cccccccc@{}}\toprule
\multirow{2}{*}{$\textrm{Variable}$} & \multirow{2}{*}{$\textrm{Unit}$} & \phantom{a}& \multicolumn{2}{c}{\textrm{Initial value}} & \phantom{a}& \multicolumn{2}{c}{\textrm{Extreme value}} \\
\cmidrule{4-5} \cmidrule{7-8}
& && $\textrm{lb}$ & $\textrm{ub}$ && $\textrm{min}$ & $\textrm{max}$ \\ \midrule
$c_A$ & \unit{\mol\per\litre} && \SI{2.14}{} & \SI{2.57}{} && \SI{1.74}{} & \SI{2.74}{} \\
$c_B$ & \unit{\mol\per\litre} && \SI{0.87}{} & \SI{1.09}{} && \SI{0.87}{} & \SI{1.28}{}\\
$T$ & \unit{\kelvin} && \SI{387}{} & \SI{403}{} && \SI{385}{}& \SI{403}{}\\
$T_K$ & \unit{\kelvin} && \SI{371}{}& \SI{386}{}&& \SI{371}{}& \SI{395}{}\\
$\frac{\Dot{V}}{V_R}$ & \unit{(1\per\hour)} && \SI{5}{} & \SI{28.4}{} && \SI{5}{} & \SI{28.4}{} \\
$\Dot{Q}_K$ & \unit{\kilo\joule\per\hour} && \SI{-2227}{} & \SI{0}{} && \SI{-2227}{} & \SI{0}{}\\
\bottomrule
\end{tabular}
\label{tab:vdv-ranges}
\end{table}

For the training, we use a hybrid strategy; we first start with the Adam optimizer \citep{Kingma2014Adam:Optimization} and then switch to the Limited-memory Broyden–Fletcher–Goldfarb–Shanno (L-BFGS) algorithm \citep{Liu1989OnOptimization}. L-BFGS typically provides more accurate results for PINNs \citep{Markidis2021TheSolvers}; however, it tends to get stuck in a local minimum if used directly \citep{Markidis2021TheSolvers}. Thus, Adam is first used to avoid local minima, and then L-BFGS is used for fine-tuning following the approach presented by \citep{Markidis2021TheSolvers, Jin2021NSFnetsEquations} since we could confirm their observation during our preliminary studies. We use a dynamic weighting scheme to decide on the weights $\lambda_i$ in Equation \eqref{eq:pinn-loss-ic-cv}, called inverse Dirichlet weighting (IDW) \citep{Maddu2022InverseNetworks}. For this purpose, we used code snippets from the GitHub repository of \cite{Maddu2022InverseNetworks}. In preliminary studies, we found IDW to yield decent results but did not perform a systematic comparison of different weighting schemes. As evidenced by the results stated below, the PINNs consistently outperform the corresponding vanilla NN benchmark models. Thus, we refrained from further investigations into different weighting schemes. As IDW only works with first-order optimizers, we apply it only during the Adam optimization step and then keep the final weights constant for the L-BFGS optimization step. We start the training process with the Adam optimizer for 1000 epochs with a learning rate of $0.001$. After that, we utilize the L-BFGS optimizer for 300 epochs. Mean squared error (MSE) is the metric used for minimization. 

To determine the architecture parameters of the PINN, we utilize a grid search varying the following hyperparameters: activation function $\in \{\textrm{tanh}, \textrm{sigmoid}\}$, depth of the hidden layers $\in \{1,2,3,4\}$ and width of hidden layers (number of nodes) $\in \{16, 32, 64, 128\}$. We investigate all four models for both data regimes. The $\textrm{tanh}$ activation function performs best in all cases. Moreover, we find that the best-performing width and depth of the hidden layers do not change across models but with the amount of training data. For the low-data regime, we find that a network with 2 hidden layers and 32 nodes performs the best. A network with 2 hidden layers and 64 nodes performs best for the high-data regime. The grid search is done with 5 randomly drawn data sets and 5 runs for each data set to account for variations in training/test split and weight initialization. Moreover, all the upcoming studies are also done using 5 data sets and 5 runs for each data set. The result of a run is reported as the average error over $N_{test} = 20$ trajectories.

\subsection{Prediction of measured states}
In this subsection, we investigate the generalization capabilities of the different PINN models and the vanilla ANN model listed in Table \ref{tab:vdv-models}. 

\begin{figure}[h]
\centering
\begin{subfigure}{.5\textwidth}
    \centering
    \includegraphics[width=1\linewidth]{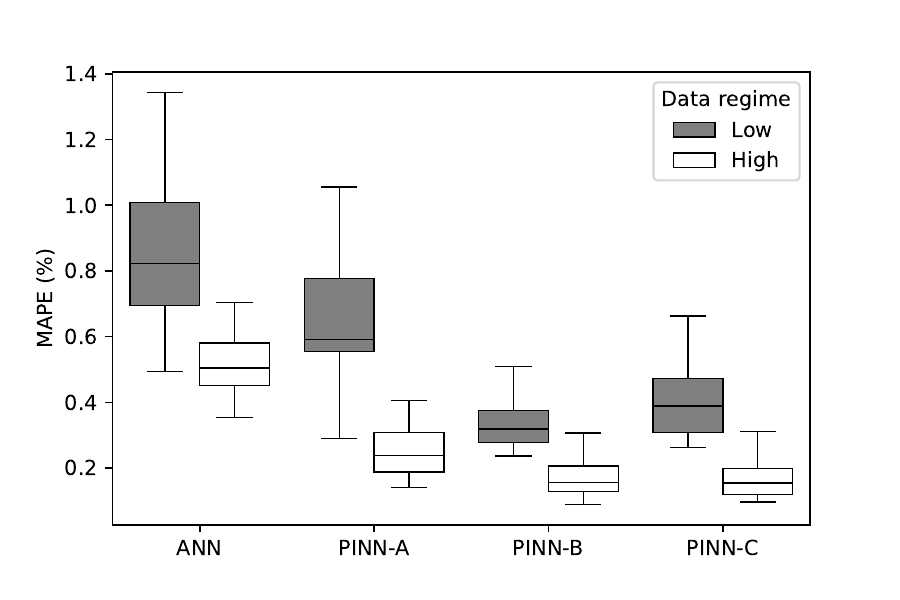}
    \caption{Concentration of reactant $c_A$.}
    \label{fig:vdv-test1-cA}
\end{subfigure}%
\begin{subfigure}{.5\textwidth}
    \centering
    \includegraphics[width=1\linewidth]{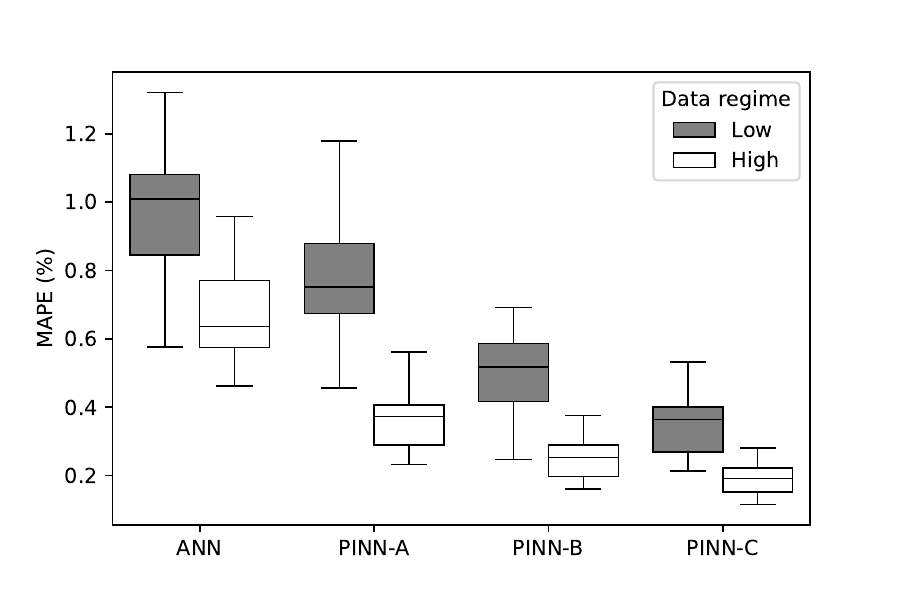}
    \caption{Concentration of product $c_B$.}
    \label{fig:vdv-test1-cB}
\end{subfigure}
\begin{subfigure}{.5\textwidth}
    \centering
    \includegraphics[width=1\linewidth]{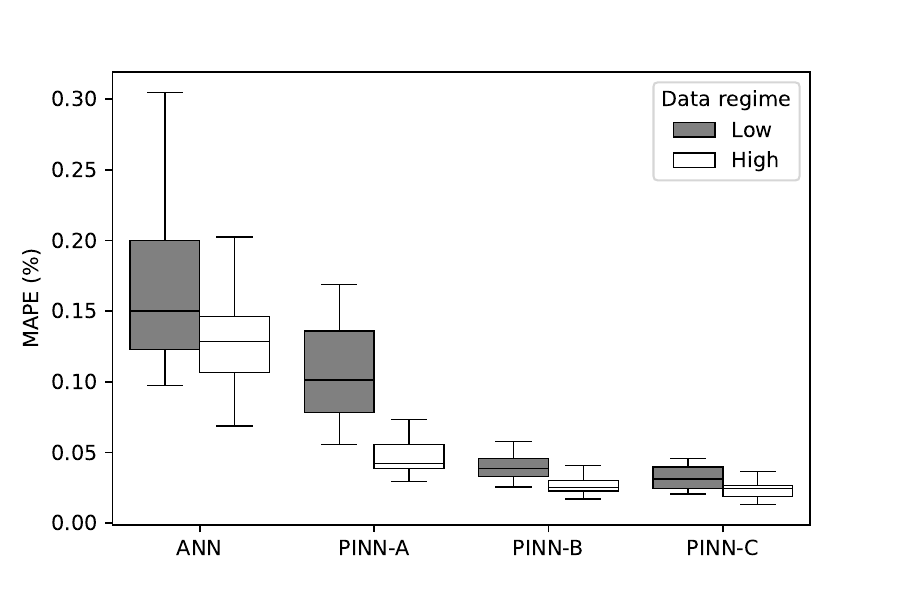}
    \caption{Temperature of the tank $T$.}
    \label{fig:vdv-test1-T}
\end{subfigure}%
\begin{subfigure}{.5\textwidth}
    \centering
    \includegraphics[width=1\linewidth]{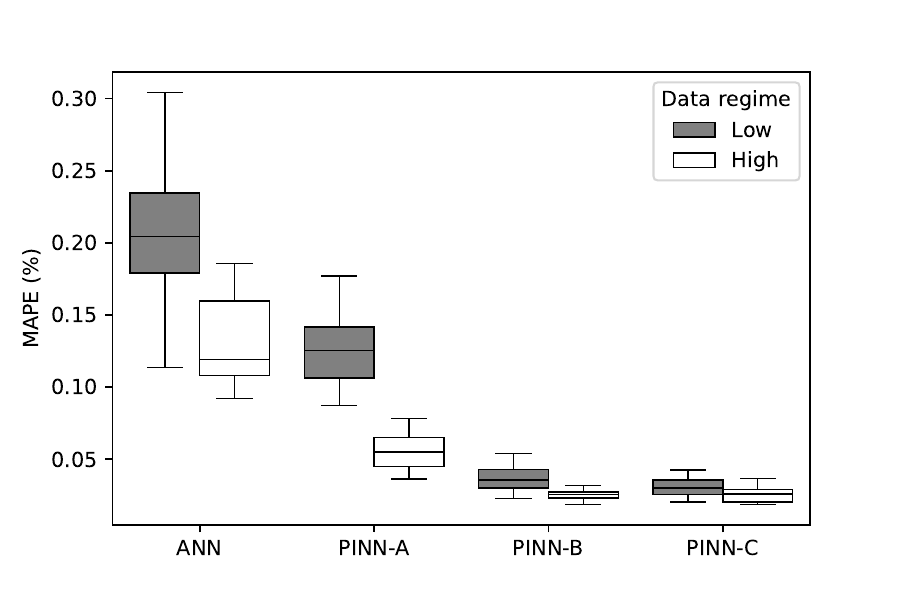}
    \caption{Cooling jacket temperature $T_K$.}
    \label{fig:vdv-test1-TK}
\end{subfigure}
\caption{Test set error for the measured states for all models and data regimes. Boxplots show the results of 25 models (5 runs each for 5 data sets), averaged over the test set of each model. The error metric is the mean absolute percentage error (MAPE).}
\label{fig:vdv-test1-1}
\end{figure}

As can be seen from Figure \ref{fig:vdv-test1-1}, the prediction error for all states decreases with increasing physical knowledge supplied to the models, except for the reactant concentration $c_A$ in the low-data regime. Moreover, all PINN models perform better than the vanilla ANN model in predicting measured states for both data regimes. A particularly interesting result is that PINN-A performs better at estimating the measured states $T$ and $T_K$ than the vanilla ANN, even though both models predict these states only based on data, i.e., PINN-A does not have energy balances, and $T$ and $T_K$ do not appear in the mole balances. A possible explanation could be that, since PINN-A has physics knowledge on $c_A$ and $c_B$, it reaches a lower loss value on $c_A$ and $c_B$ than the vanilla ANN and thus has more room to optimize for $T$ and $T_K$. 

We conclude that the PINN models show strong generalization capabilities, better than the purely data-driven model, especially in the low-data regime.

\subsection{Algebraic state estimation}
\label{study-1}
We now investigate if the PINN models can predict unmeasured algebraic states $\Vy^u$ with reasonable accuracy. First, we conduct an incidence matrix analysis for each PINN model.  Table \ref{tab:struc-matr-vdv} shows that all PINN models have a full-column rank incidence matrix, suggesting that state estimation is possible in all cases.   

\begin{table}[h]
    \centering
    \caption{Incidence matrices of PINN-A, PINN-B, and PINN-C for the Van de Vusse reactor example. If an unmeasured state appears in an equation, it is marked with a cross. Encircled crosses show feasible assignments of states to equations.}
        \begin{subtable}[c]{\textwidth}
        \centering
        \caption{Incidence matrix of PINN-A. Matrix has full-column rank.}
        \label{tab:ca-m1}
        \begin{tabular}{ l|c|c }
         $[\bm f, \bm g]\downarrow \quad [\Vx^u, \Vy^u]\rightarrow$        & $r_A$ & $r_B$ \\ \hline
        Eqn. for $\Dot{c}_A$ & $\otimes$ &    \\ \hline
        Eqn. for $\Dot{c}_B$ &    & $\otimes$  
        \end{tabular}
        \end{subtable}\\        
        \vspace{6mm}
        \begin{subtable}[c]{\textwidth}
        \centering
        \caption{Incidence matrix of PINN-B. Matrix has full-column rank.}
        \label{tab:ca-m2}
        \begin{tabular}{ l|c|c|c }
         $[\bm f, \bm g]\downarrow \quad [\Vx^u, \Vy^u]\rightarrow$         & $r_1$ & $r_2$ & $r_3$ \\ \hline
        Eqn. for $\Dot{c}_A$ & $\otimes$  &    & $\times$  \\ \hline
        Eqn. for $\Dot{c}_B$ & $\times$  & $\otimes$  &    \\ \hline
        Eqn. for $\Dot{T}$ & $\times$  & $\times$  & $\otimes$  \\ \hline
        Eqn. for $\Dot{T}_K$ &    &    &   
        \end{tabular}
        \end{subtable}\\  
        \vspace{6mm}
        \begin{subtable}[c]{\textwidth}
        \centering
        \caption{Incidence matrix of PINN-C. Matrix has full-column rank.}
        \label{tab:ca-m3}
        \begin{tabular}{ l|c|c|c }
         $[\bm f, \bm g]\downarrow \quad [\Vx^u, \Vy^u]\rightarrow$        & $k_1$ & $k_2$ & $k_3$ \\ \hline
        Eqn. for $\Dot{c}_A$ & $\otimes$  &    & $\times$  \\ \hline
        Eqn. for $\Dot{c}_B$ & $\times$  & $\otimes$  &    \\ \hline
        Eqn. for $\Dot{T}$ & $\times$  & $\times$  & $\otimes$  \\ \hline
        Eqn. for $\Dot{T}_K$ &    &    &  
        \end{tabular}
        \end{subtable}

    \label{tab:struc-matr-vdv}
\end{table}

Table \ref{tab:test1-unmeasured states} reports test errors for the unmeasured algebraic states. Since the compared quantities are different for the different models, a direct comparison between the models is not justified. However, we can conclude that all models can predict the unmeasured algebraic states with acceptable accuracy (less than 10 \% mean absolute percentage error), except $r_3$ in PINN-B. We also observe that the accuracy gap between the low and high data regimes decreases as the provided physics knowledge increases.

Note that the estimated algebraic states, i.e., the net reaction rates (PINN-A), the individual reaction rates (PINN-B), and the reaction rate constants (PINN-C), were not only unmeasured, i.e., no process data was used for training, but also their corresponding constitutive equations were not provided. This example thus shows that PINNs can, in certain situations, infer immeasurable states, even if respective constitutive equations are unknown. 

\begin{table}[h]
\centering
\caption{Estimation accuracy for the unmeasured algebraic states $\Vy^u$ on the test set for all PINN models and data regimes. Results are averaged over 25 models (5 runs each for 5 data sets). The error metric is the mean absolute percentage error (MAPE ).}
\begin{tabular}{@{}ccccc@{}}\toprule 
{$\textrm{Model}$} & {Unmeasured algebraic state} & {Low data regime} & {High data regime}\\ \midrule
\multirow{2}{*}{$\textrm{PINN-A}$} & $r_A$ & 4.71\% & 2.61\%\\
&$r_B$ & 9.31\% & 5.12\% \\
\midrule
\multirow{3}{*}{$\textrm{PINN-B}$} & $r_1$ & 4.33\%  & 3.43\% \\
&$r_2$ & 9.15\% & 7.27\% \\
&$r_3$ & 11.99\% & 10.42\%\\
\midrule
\multirow{3}{*}{$\textrm{PINN-C}$} & $k_1$ & 3.59\%  &  2.90\% \\
&$k_2$ & 6.84\% & 6.13\%\\
&$k_3$ & 7.14\%& 6.98\%\\
\bottomrule
\end{tabular}
\label{tab:test1-unmeasured states}
\end{table}

\subsection{Differential state estimation}
\label{sec:diff-se}
We study PINN-C and create three different settings to empirically gauge the differential state estimation capabilities. In the first setting, we assume that state $c_A$ is unmeasured i.e., $\Vx^u = [c_A]^T$. In the second setting, $T$ is unmeasured, i.e., $\Vx^u = [T]^T$. In the third setting, $T_K$ is unmeasured, i.e., $\Vx^u = [T_K]^T$. For all settings, the algebraic states $k_1$, $k_2$, and $k_3$ are also unmeasured, i.e., $\Vy^u = [k_1, k_2, k_3]^T$.

In the first setting, $\Vx^u = [c_A]^T$, we do not obtain a full-column rank incidence matrix, as can be seen from Table \ref{tab:s3-m3-ca}, whereas in the other two settings we do (Tables \ref{tab:s3-m3-T} and \ref{tab:s3-m3-TK}).

\begin{table}[h]
    \centering
        \caption{Incidence matrices of PINN-C with $\Vx^u = [c_A]^T$, $\Vx^u = [T]^T$ and $\Vx^u = [T_K]^T$ for Van de Vusse reactor example. If an unmeasured state appears in an equation, it is marked with a cross. Encircled crosses show feasible assignments of states to equations.}
        \begin{subtable}[c]{\textwidth}
        \centering
        \caption{Incidence matrix of PINN-C with $\Vx^u = [c_A]^T$ (setting 1). Matrix does \emph{not} have full-column rank.}
        \label{tab:s3-m3-ca}
        \begin{tabular}{l|c|c|c|c}
        $[\bm f, \bm g]\downarrow \quad [\Vx^u, \Vy^u]\rightarrow$         & $c_A$ & $k_1$ & $k_2$ & $k_3$ \\ \hline
        Eqn. for $\Dot{c}_A$ & $\times$  & $\times$  &    & $\times$  \\ \hline
        Eqn. for $\Dot{c}_B$ & $\times$   & $\times$  & $\times$  &    \\ \hline
        Eqn. for $\Dot{T}$ & $\times$
        & $\times$  & $\times$  & $\times$  \\ \hline
        Eqn. for $\Dot{T}_K$ & &    &    &  
        \end{tabular}
        \end{subtable}\\
        \vspace{6mm}
        \begin{subtable}[c]{\textwidth}
        \centering
        \caption{Incidence matrix of PINN-C with $\Vx^u = [T]^T$ (setting 2). Matrix has full-column rank.}
        \label{tab:s3-m3-T}
        \begin{tabular}{ l|c|c|c|c }
         $[\bm f, \bm g]\downarrow \quad [\Vx^u, \Vy^u]\rightarrow$        & $T$ & $k_1$ & $k_2$ & $k_3$ \\ \hline
        Eqn. for $\Dot{c}_A$ &   & $\otimes$  &    & $\times$  \\ \hline
        Eqn. for $\Dot{c}_B$ &    & $\times$  & $\otimes$  &    \\ \hline
        Eqn. for $\Dot{T}$ & $\times$
        & $\times$  & $\times$  & $\otimes$  \\ \hline
        Eqn. for $\Dot{T}_K$ & $\otimes$ &    &    &   
        \end{tabular}
        \end{subtable}\\
        \vspace{6mm}
        \begin{subtable}[c]{\textwidth}
        \centering
        \caption{Incidence matrix of PINN-C with $\Vx^u = [T_K]^T$ (setting 3). Matrix has full-column rank.}
        \label{tab:s3-m3-TK}
        \begin{tabular}{ l|c|c|c|c }
        \hline
        $[\bm f, \bm g]\downarrow \quad [\Vx^u, \Vy^u]\rightarrow$           & $T_K$ & $k_1$ & $k_2$ & $k_3$ \\ \hline
        Eqn. for $\Dot{c}_A$ &    & $\otimes$  &    & $\times$  \\ \hline
        Eqn. for $\Dot{c}_B$ &    & $\times$  & $\otimes$  &    \\ \hline
        Eqn. for $\Dot{T}$ & $\times$  & $\times$  & $\times$  & $\otimes$  \\ \hline
        Eqn. for $\Dot{T}_K$ & 
        $\otimes$
        &    &    & 
        \end{tabular}
        \end{subtable}
    \label{tab:struc-matrix-vdv-test3}
\end{table}

\begin{figure}[h]
    \centering
    \includegraphics[width=1\linewidth]{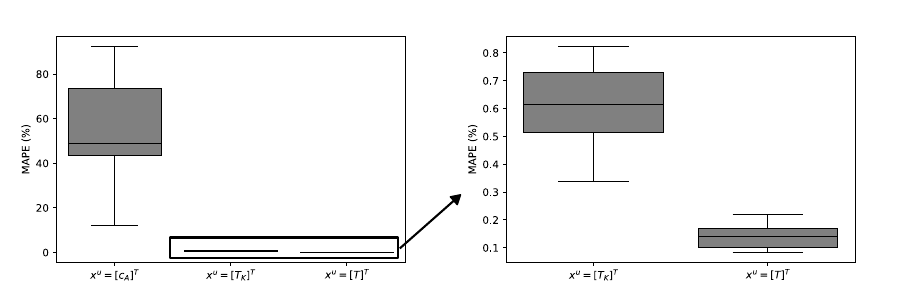}
    \caption{Test set errors for the unmeasured differential states of PINN-C with $\Vx^u = [c_A]^T$, $\Vx^u = [T]^T$ and $\Vx^u = [T_K]^T$ for the Van de Vusse reactor example. All error values correspond to the respective unmeasured differential state, e.g., the value for the model with $\Vx^u = [c_A]^T$ shows the error of $c_A$. Boxplots show the results of 25 models (5 runs each for 5 data sets), averaged over the test set of each model. The error metric is the mean absolute percentage error (MAPE).}
    \label{fig:vdv-test3-mape}
\end{figure}

\begin{figure}[h] 
    \centering\captionsetup{width=.8\linewidth}
        \includegraphics[width=.5\linewidth]{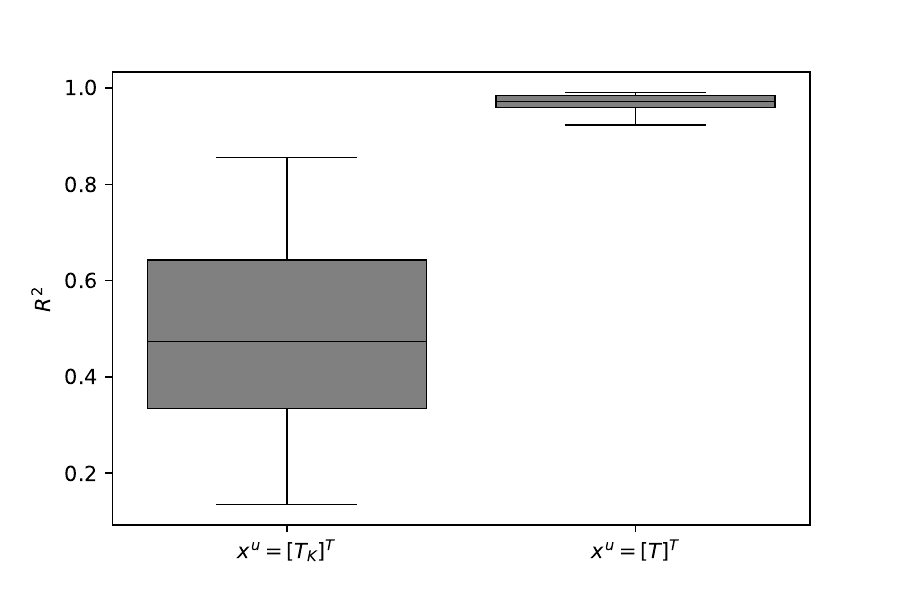}
        \caption{$R^2$ values for PINN-C with $\Vx^u = [T_K]^T$ and $\Vx^u = [T]^T$ (test set goodness of fit).}
        \label{fig:vdv-test3-R2}
\end{figure}

In Figure \ref{fig:vdv-test3-mape}, we see that the PINN model with $\Vx^u = [c_A]^T$ (setting 1) indeed fails to estimate $c_A$, as indicated by the incidence matrix (Table \ref{tab:s3-m3-ca}). In contrast, the MAPE values suggest that the PINN models with $\Vx^u = [T]^T$ (setting 2) and $\Vx^u = [T_K]^T$ (setting 3) yield good results for the estimation of the respective unmeasured differential states $T$ and $T_K$. However, when we compare the results to the case where all differential states were measured (cf. Figure \ref{fig:vdv-test1-1}), we see that the MAPE values are about 20 times higher in case of $T_K$, and around 5 times higher in case of $T$. More importantly, as the ranges of $T$ and $T_K$ are quite low compared to the actual values (cf. Table \ref{tab:vdv-ranges}), the MAPE values can be deceptively low. Thus, as a more reliable measure of goodness of fit, we evaluate the coefficient of determination ($\mathrm{R^2}$). As can be seen from Figure \ref{fig:vdv-test3-R2}, the PINN-C model with $\Vx^u = [T]^T$ can successfully predict the unmeasured differential state $T$, with $\mathrm{R^2}$ scores above 0.90.
However, the PINN-C model with $\Vx^u = [T_K]^T$ essentially fails to estimate $T_K$, with $\mathrm{R^2}$ scores ranging between 0.15 and 0.85. 

In state estimation theory \citep{kalman1960general,lee1967foundations}, a system is called \emph{observable} if the initial values of unmeasured states can be estimated uniquely using the information on measured states and a mathematical process model. In the particular example considered here, the initial state (and thus the trajectory) of $T$ can be estimated uniquely by the PINN using the data on the measured states and the built-in physical knowledge, whereas this is not the case for $T_K$. Transferring observability conditions for nonlinear dynamic models, see, e.g., \cite{lee1967foundations,kou1973observability}, to PINNs is not straightforward and thus considered beyond the scope of this paper. 

\subsection{Extrapolation capabilities}
We now explore if the PINN can extrapolate beyond the bounds of the process data supplied for training. For this purpose, we create a set of test trajectories with the initial value of $c_{A0}$ out of the bounds of $c_{A0}$ in the training trajectories. We term this set \textit{extrapolation set}. In Table \ref{tab:vdv-u-ranges}, respective ranges for the inputs $c_{A0}$ can be seen for training, test, and extrapolation sets.
In contrast to purely data-driven models, the PINNs may also learn the system dynamics from the physics residuals.
Nevertheless, we still expect a lower accuracy in the extrapolation regime since we do not provide measurement data about that regime during training.

\begin{figure}[h]
    \centering
    \includegraphics[width=.5\linewidth]{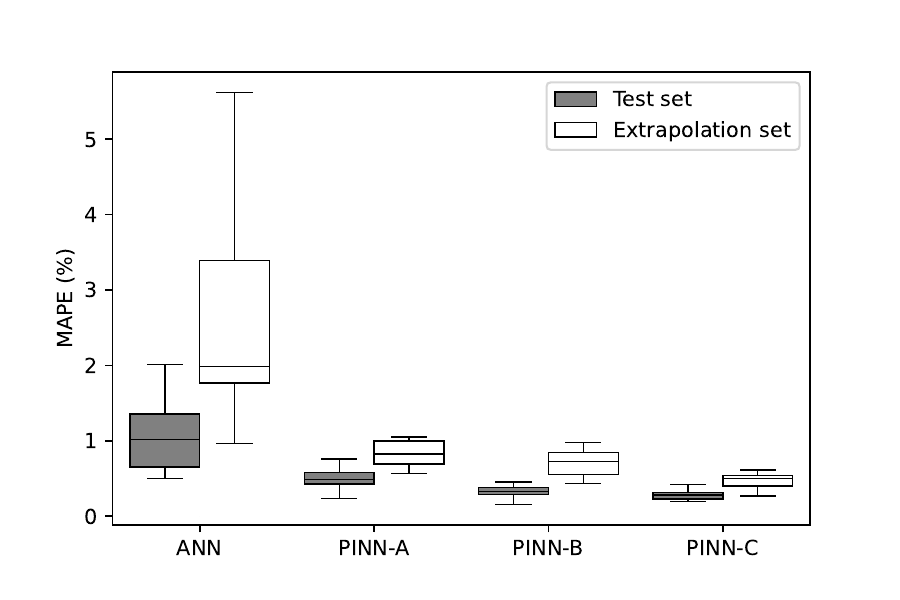}
    \caption{Test and extrapolation set errors of the reactant concentration $c_A$ for all models. The results are on the low-data regime. The error metric is the mean absolute percentage error (MAPE).}
    \label{fig:vdv-test4-cA}
\end{figure}

As can be seen from Figure \ref{fig:vdv-test4-cA}, the test errors on both the test and the extrapolation sets are much lower for the PINN models compared to the vanilla ANN model. We also observe that PINN-C, the model with the most physics knowledge, has the lowest difference in accuracy between test and extrapolation sets. Thus, we conclude that the PINN models can extrapolate better than the non-informed NN, and the extrapolation accuracies tend to increase when more physics knowledge is incorporated into the PINN. 

\begin{table}[h]
\caption{Ranges of the initial state $c_{A0}$ for the trajectories used in the training, test, and extrapolation sets.}
\centering
\begin{tabular}{@{}cccc@{}}
\toprule 
Set & Initial State & Lower bound & Upper bound\\
\midrule
Training set & $c_{A0}$ & $2.14 \frac{mol}{L}$  & $2.57 \frac{mol}{L}$ \\
Test set & $c_{A0}$ & $2.14 \frac{mol}{L}$  & $2.57 \frac{mol}{L}$  \\
Extrapolation set & $c_{A0}$ & $1.71 \frac{mol}{L}$  & $2.14 \frac{mol}{L}$  \\
\bottomrule
\end{tabular}
\label{tab:vdv-u-ranges}
\end{table}

\section{Numerical example 2: Liquid-liquid separator}\label{sec:Settler}
With this second example, we aim to investigate the generalization and state estimation capabilities of the PINN models under varying amounts of measured physical property data supplied as additional inputs to the NN. Thus, we now focus on the data-driven part of the PINN.

The dynamic liquid-liquid separator model shown below is based on the work of \cite{Backi2018AFirst-Principles, Backi.2019}. We included extensions for swarm sedimentation in the aqueous phase, convection terms for the drop size distribution (DSD) in the dense-packed zone (DPZ) analogously to \cite{Backi2018AFirst-Principles}, and a state-of-the-art coalescence model \citep{Henschke1995DimensionierungAbsetzversuche}. The chosen swarm model \citep{Mersmann.1980} was also used to model liquid-liquid columns \citep{Kampwerth2020TowardsModel} and takes the form of Stokes' law \citep{Stokes1850OnIII} for diminishing hold-ups. Stokes' law was experimentally confirmed to model the outlet hold-up of liquid-liquid separator accurately \citep{Ye2023ImpactEfficiency}. 

\begin{figure}[h]
    \centering
    \includegraphics[scale=0.5]{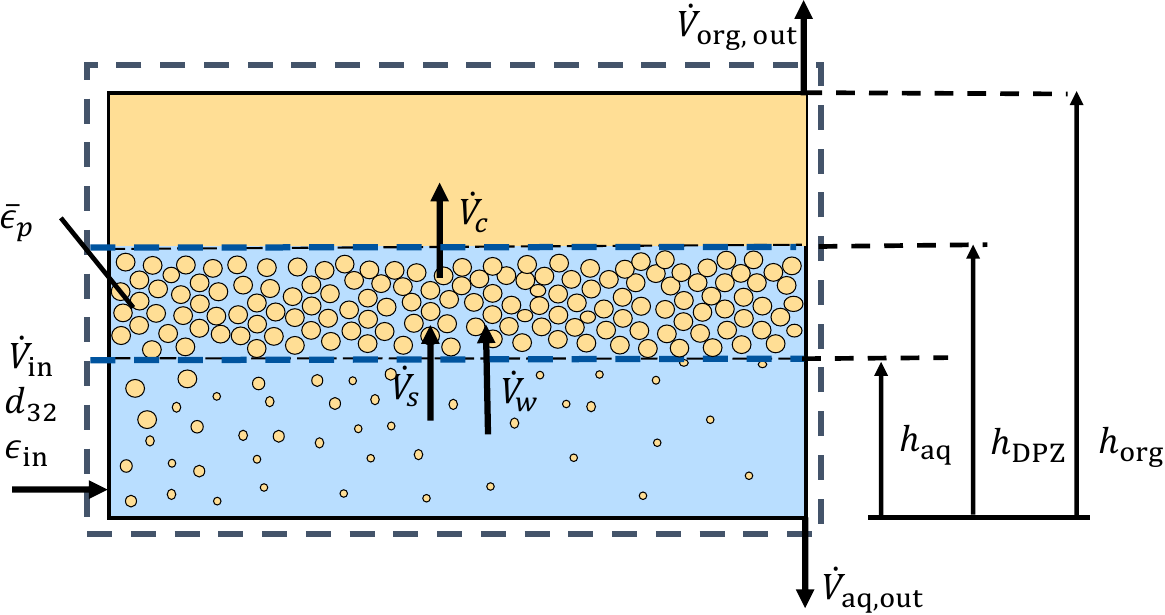}
    \caption{Separator with the light phase (top), dense-packed zone (center), heavy phase (bottom), and flows. The dispersion with the properties phase fraction of dispersed phase $\epsilon_\textrm{in}$, Sauter mean diameter $d_{32}$, and total volume flow rate $\dot{V}_\textrm{in}$ enters the heavy phase from the left. The heavy phase has the following outgoing flows: sedimentation rate $\dot{V}_s$, water flow rate $\dot{V}_w$ and outlet flow $\dot{V}_{\textrm{aq,out}}$. The dense-packed zone is modeled with a constant hold-up $\bar{\epsilon}_p = 0.9$ and a coalescence rate $\dot{V}_c$. The light phase has the outlet $\dot{V}_{\textrm{org,out}}$.}
    \label{fig:0D_model}
\end{figure}

The considered liquid-liquid separator shown in Fig.~\ref{fig:0D_model} is divided into three subsystems: light (organic) phase, dense-packed zone (DPZ), and heavy (aqueous) phase. The light phase is assumed to be free of the dispersed phase; the DPZ is assumed to have a constant hold-up $\bar{\epsilon}_p$ (volume phase fraction of dispersed phase) of \SI{0.9}{}, and the heavy phase contains dispersed organic droplets but mostly water. The total volume flow $\dot{V}_{\textrm{in}}$ enters the separator in the heavy phase with the dispersed light phase described by the Sauter mean diameter $d_{32}$ and phase fraction of organic phase $\epsilon_{\textrm{in}}$. In the heavy phase, the dispersed droplets sediment upwards as a droplet swarm, resulting in the volume flow of organic droplets to the DPZ $\dot{V}_s$. In the DPZ, drop-drop coalescence is assumed to be negligible, and only droplet-interface coalescence occurs, giving the volume flow of coalesced drops $\dot{V}_c$ to the light phase. The volume flow of water $\dot{V}_w$ from the aqueous phase to the DPZ stems from trapped water between the sedimented droplets and coalesced drops at the interface of the organic phase. By applying a volume balance to the DPZ and assuming a constant hold-up, the volume flow of water can be expressed by the sedimentation and coalescence rate. The outlet volume flow of the aqueous  $\dot{V}_\textrm{aq,out}$ and organic phase $\dot{V}_\textrm{org,out}$ are the manipulated variables of the settler, $\Vu = [\dot{V}_\textrm{aq,out}, \dot{V}_\textrm{org,out}]^T$.

The following volume balance equations are obtained after transforming the volume of a cylindrical segment to the height of each segment \citep{Backi2018AFirst-Principles}:
\begin{subequations}
    \label{eq:settler}
    \begin{align}
        \label{eq:settler-1}\Dot{h}_L(t) &= \frac{\dot{V}_{in}(t) - \Dot{V}_{aq,out}(t) - \Dot{V}_{org,out}(t) }{2L\sqrt{h_L(t)(2r-h_L(t))}}, \\
        \label{eq:settler-2}\Dot{h}_{\textrm{DPZ}}(t) &= \frac{\dot{V}_{in}(t) - \Dot{V}_{aq,out}(t) - \Dot{V}_c(t) }{2L\sqrt{h_{\textrm{DPZ}}(t)(2r-h_{\textrm{DPZ}}(t))}}, \\
        \label{eq:settler-3}\Dot{h}_{\textrm{aq}}(t) &= \frac{\dot{V}_{in}(t) - \Dot{V}_{aq,out}(t) - \Dot{V}_s(t) \frac{1}{\bar{\epsilon}_p} +  \Dot{V}_c(t) \frac{1-\bar{\epsilon}_p}{\bar{\epsilon}_p} }{2L\sqrt{h_{\textrm{aq}}(t)(2r-h_{\textrm{aq}}(t))}}
    \end{align}
\end{subequations}   
Here, $h_L$, $h_{\textrm{DPZ}}$, and $h_{\textrm{aq}}$ are the heights of the total liquid, the DPZ, and the aqueous phase, respectively, each measured from the bottom of the separator. They constitute the differential states $\Vx$ of the system. Note that the volume flow rates $\dot{V}_{in}$, $\dot{V}_{aq,out}$, $\dot{V}_{org,out}$, $\dot{V}_c$, and $\dot{V}_s$ are algebraic quantities. Similar to the CSTR case (Section \ref{sec:results}), use of the dot notation to indicate flow rates is motivated by standard practice in engineering. In contrast, the dot symbols on the LHS of Equations \eqref{eq:settler-1}-\eqref{eq:settler-3} denote derivatives with respect to time. 
In the full-order mechanistic model (see Section SM4 of the Supporting Materials), the coalescence and sedimentation rates $\dot{V}_c$ and $\dot{V}_s$ are functions of $h_{\textrm{aq}}$ and $h_{\textrm{DPZ}}$, boundary conditions at the entrance and physical properties such as the Sauter mean diameter $d_{32}$ and the coalescence parameter $r_{\textrm{v}}$.
As they cannot be measured, we aim to estimate $\dot{V}_c$ and $\dot{V}_s$ with a PINN model that uses only Equations \eqref{eq:settler-1}-\eqref{eq:settler-3} as available physical knowledge, i.e., the constitutive equations for the coalescence and sedimentation rates $\dot{V}_c$ and $\dot{V}_s$ are assumed to be unknown. 

We assume that the total liquid height in the separator is constant, as this is the usual mode of operation. Then, the differential balance Equation \eqref{eq:settler-1} becomes an algebraic relation, serving as a closure condition for the flows in and out of the separator:
\begin{equation*}
    \dot{V}_{\textrm{in}}(t) - \Dot{V}_{\textrm{aq,out}}(t) - \Dot{V}_{\textrm{org,out}}(t) = 0
\end{equation*}
We also aim to investigate whether the PINN can take advantage of measurement data on $d_{32}$ and $r_{\textrm{v}}$ that are provided as input to the NN although these quantities do not appear in the physics Equations \eqref{eq:settler-1}-\eqref{eq:settler-3}.  Thus, we create three different PINN models with an increasing number of physical properties added as inputs to the NN, along with a vanilla NN for comparison (cf. Table \ref{tab:settler-models}). Moreover, we show the network structure of the models in the Supplementary Materials. As in Section \ref{sec:results}, we make the time, states, and manipulated variables dimensionless for better performance and stability during NN training. We give the dimensionless variables and equations in the Supplementary Materials.

\begin{table}[h]
\centering
\caption{Inputs of the models for the liquid-liquid separator.}
\label{tab:settler-models}
\begin{tabular}{ll} \toprule
Model name & Network inputs \\ \midrule
Vanilla ANN & $t, h_{\textrm{aq}}(t_0), h_{\textrm{DPZ}}(t_0), \Dot{V}_{\textrm{aq,out}}, \Dot{V}_{\textrm{org,out}}$   \\ 
Base PINN       &              $t, h_{\textrm{aq}}(t_0), h_{\textrm{DPZ}}(t_0), \Dot{V}_{\textrm{aq,out}}, \Dot{V}_{\textrm{org,out}}$   \\ 
PINN-$\textrm{d}_\textrm{32}$   &        $t, h_{\textrm{aq}}(t_0), h_{\textrm{DPZ}}(t_0), \Dot{V}_{\textrm{aq,out}}, \Dot{V}_{\textrm{org,out}}, d_{32}$ \\ 
PINN-$\textrm{d}_\textrm{32}$-$\textrm{r}_\textrm{v}$     &     $t, h_{\textrm{aq}}(t_0), h_{\textrm{DPZ}}(t_0), \Dot{V}_{\textrm{aq,out}}, \Dot{V}_{\textrm{org,out}},d_{32}, r_{\textrm{v}}$           \\
\bottomrule
\end{tabular}
\end{table}

\subsection{Data set generation, training, and hyperparameter selection}
We investigate the phase separation of n-butyl acetate dispersed in water in a pilot-scale separator. The radius and length of the separator are $R = \SI{0.1}{\meter}$ and $L = \SI{1.8}{\meter}$. We take the operating ranges presented in Table \ref{tab:settler-ranges}, with a selected time interval for step-wise control changes and thus process time of $20 \thinspace \textrm{s}$, i.e., $t \in [0, 20] \thinspace \textrm{s}$. 
We keep the manipulated variables constant throughout the process time for implementation reasons, as done in Section \ref{sec:results}.  
Data generation to calculate the physics loss term $MSE_{physics}$ and the initial condition loss term $MSE_{init}$ in Equations \eqref{eq:pinn-loss-ic-cv} are done by selecting $N_e = 10000$ collocation and $N_i=100$ initial value points. Again, the selection is done using Latin Hypercube sampling.  We choose the bounds for the initial states $\Vx(t_0)$ corresponding to the minimum and maximum values of the states $\Vx$ in the operating range of the process (see Table \ref{tab:settler-ranges}), and perform similarly for the control variables $\Vu$. 
We use the explicit Runge-Kutta method of order 5 for the process data generation, utilizing \textit{solve\_ivp} solver from \textit{scipy.integrate} module in Python \citep{Virtanen2020SciPyPython,Dormand1980AFormulae}, with a relative and absolute error of \num{1e-12}. We output the trajectories on a time grid $t \in [0, 20] \thinspace \textrm{s}$ with $\Delta t = 0.1 \thinspace \textrm{s}$. Nonphysical states, such as flooding of the separator with the DPZ, are addressed by early termination. The resulting shorter trajectories are kept in the data set; however, the step size $\Delta t$ is adjusted to keep a constant number of grid points among all trajectories. 
We create $N_{total} = 200$ trajectories. From these, we select $N_{test} = 40$ trajectories for testing. For training, we use $N_{train}$ trajectories, each having $N_m = 201$ data points. The total number of measurement points are $N_d = N_{train} N_m$. Again, we create two training sets from the remaining 160 trajectories not used for testing: a training set representing a \textit{low-data regime} consisting of only $N_{train} = 20$ training trajectories, and a training set representing a \textit{high-data regime} consisting of $N_{train} = 160$ training trajectories. 

\begin{table}[h]
\centering
\ra{1.3}
\caption{Ranges for initial states and inputs for the liquid-liquid separator example.}
\begin{tabular}{lll}\toprule
Variable & Lower bound & Upper bound \\
\midrule
$h_{\textrm{aq},0}$ & \SI{0.090}{\meter}& \SI{0.110}{\meter}  \\
$h_{\textrm{DPZ},0}$ & \SI{0.108}{\meter}& \SI{0.132}{\meter}\\
$\Dot{V}_{\textrm{aq,out}}$ & \SI{4.5e-4}{\cubic\meter\per\second}& \SI{5.5e-4}{\cubic\meter\per\second}\\
$\Dot{V}_{\textrm{org,out}}$ & \SI{2.0e-4}{\cubic\meter\per\second}& \SI{5.0e-4}{\cubic\meter\per\second} \\
$d_{32}$ &  \SI{9.0e-4}{\meter}& \SI{1.1e-3}{\meter}\\
$r_{\textrm{v}}$ & 0.033 & 0.043\\
\bottomrule
\end{tabular}
\label{tab:settler-ranges}
\end{table}

We use the strategy described in Section \ref{sec:vdv-data} for the training and hyperparameter optimization. For the low-data regime, we find that a network with two hidden layers and 32 nodes performs the best. A network with two hidden layers and 128 nodes performs best for the high-data regime. The $\textrm{tanh}$ activation function performs best in all cases. The grid search is done with 5 data sets and 5 runs for each data set to account for variations in training/test split and weight initialization. Moreover, we use a sigmoid activation function for the output layer to bound the output values between 0 and 1 to prevent the square root in the denominator of Equations \eqref{eq:settler-2} and \eqref{eq:settler-3} from attaining negative values during PINN training. The following numerical studies are done with 5 data sets and 5 runs for each data set. The results of the runs are reported as the average error over $N_{test} = 20$ trajectories.

\subsection{Results}
\label{sec:settler-results}
We show the incidence matrix of the liquid-liquid separator PINN model in Table \ref{tab:settler-strmat}. The unmeasured NN outputs are the algebraic states ${\Vy}^u = [\Dot{V}_c, \Dot{V}_s]$. The total liquid height is known and constant, i.e., $\Dot{h}_L = 0$. The incidence matrix shows a feasible assignment and thus indicates possible state estimation.

\begin{table}[h]
\centering
\caption{Incidence matrix of the PINN models for the liquid-liquid separator. The matrix is identical for all three PINN models. If an unmeasured state appears in an equation, it is marked with a cross. Encircled crosses show the feasible assignment of states to equations. The matrix has full column rank.}
\label{tab:settler-strmat}
\begin{tabular}{ l|c|c }
    $[\bm f, \bm g]\downarrow \quad [\Vx^u, \Vy^u]\rightarrow$       & $\Dot{V}_c$ & $\Dot{V}_s$ \\ \hline
Eqn. \eqref{eq:settler-1} &  &   \\ \hline
Eqn. \eqref{eq:settler-2} & $\otimes$ &   \\ \hline
Eqn. \eqref{eq:settler-3} &  $\times$  & $\otimes$ 
\end{tabular}
\end{table}

\begin{figure}[h]
\centering
\begin{subfigure}{.5\textwidth}
    \centering
    \includegraphics[width=1\linewidth]{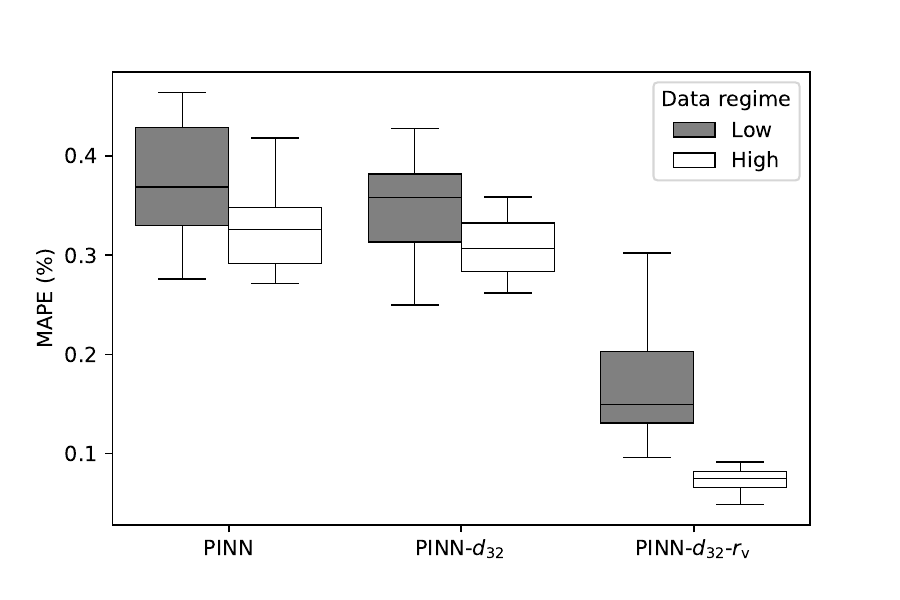}
    \caption{Dense-packed zone height $h_{\textrm{DPZ}}$.}
    \label{fig:0d-settler-hdpz_wo_NN}
\end{subfigure}%
\begin{subfigure}{.5\textwidth}
    \centering
    \includegraphics[width=1\linewidth]{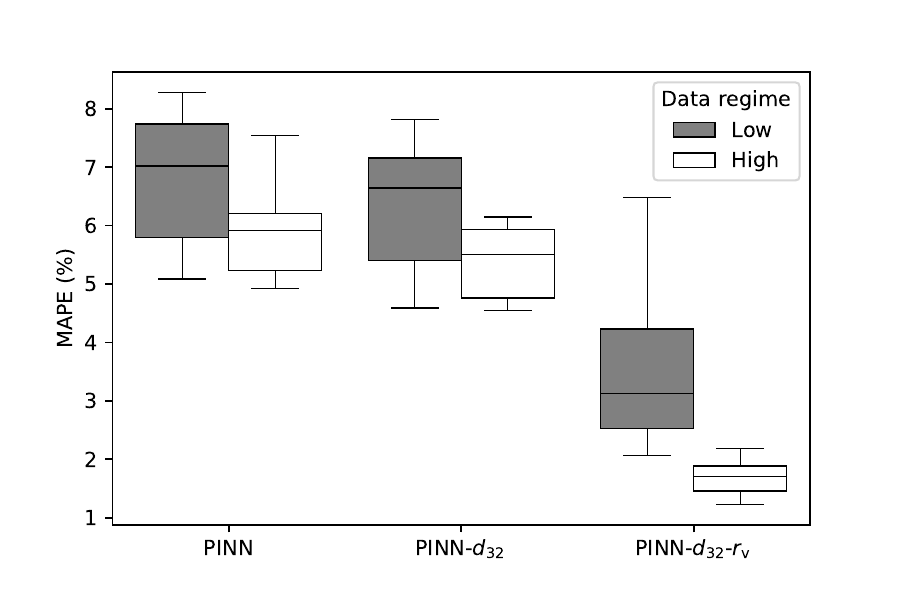}
    \caption{Coalescence rate $\Dot{V}_c$.}
    \label{fig:0d-settler-qc_wo_NN}
\end{subfigure}
\caption{Test set error for the DPZ height $h_{\textrm{DPZ}}$ and the coalescence rate $\Dot{V}_c$ for all PINN models and data regimes. The error metric is the mean absolute percentage error (MAPE).}
\label{fig:settler-results-1}
\end{figure}

We compare the prediction error of the states of the liquid-liquid separator model based on test set data. As can be seen from Figure \ref{fig:0d-settler-hdpz_wo_NN}, the prediction accuracy of the DPZ height $h_{\textrm{DPZ}}$ increases slightly with the addition of the Sauter mean diameter at the inlet $d_{32}$ as a NN input. However, a more drastic increase can be noted if the coalescence parameter $r_{\textrm{v}}$ is added as NN input. We see a similar trend with the estimation of the coalescence rate $\Dot{V}_c$ in Figure  \ref{fig:0d-settler-qc_wo_NN}, although the accuracy increase is more apparent in the high-data regime. As explained in the Supplementary Materials, the coalescence parameter $r_{\textrm{v}}$ plays a more direct role in the determination of the coalescence rate $\Dot{V}_c$, which in turn has a high impact on $h_{\textrm{DPZ}}$ (Equation \eqref{eq:settler-2}). The Sauter mean diameter at the inlet $d_{32}$ has only an indirect role since the sub-model for coalescence and sedimentation in the full-order mechanistic model (see Section SM4 of the Supporting Materials) divides the separator into segments through the axial length. Thus, the Sauter mean diameter at each segment $d_{32,i}$ determines the coalescence rate rather than the value at the inlet. Moreover, since the PINN models are not trained with the data of the coalescence rate $\Dot{V}_c$, and the sub-model for the coalescence rate is not provided as physics knowledge, the estimation accuracy of the coalescence rate $\Dot{V}_c$ highly depends on the prediction accuracy of $h_{\textrm{DPZ}}$. 

\begin{figure}[h]
\centering
\begin{subfigure}{.5\textwidth}
    \centering
    \includegraphics[width=\linewidth]{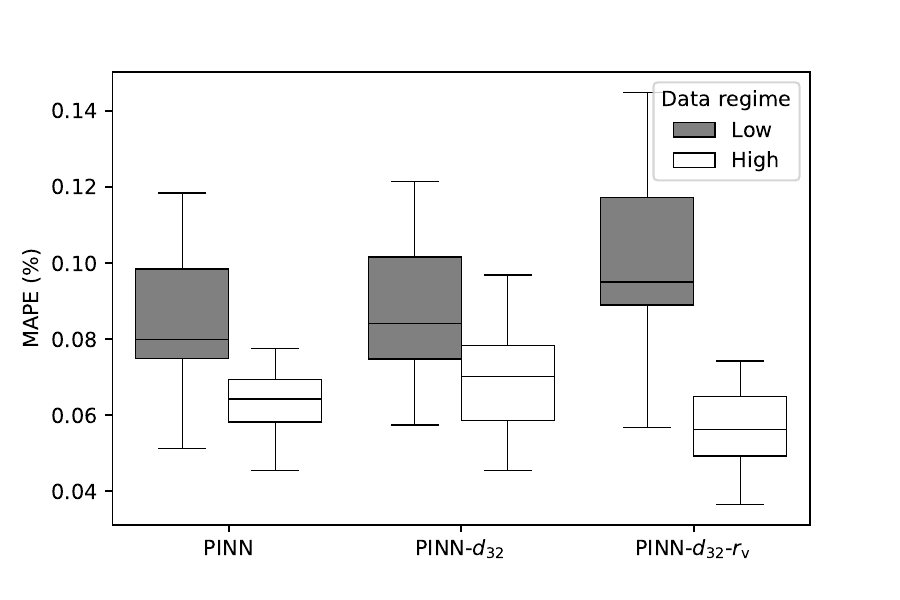}
    \caption{Water height $h_{\textrm{aq}}$.}
    \label{fig:0d-settler-haq_wo_NN}
\end{subfigure}%
\begin{subfigure}{.5\textwidth}
    \centering
    \includegraphics[width=\linewidth]{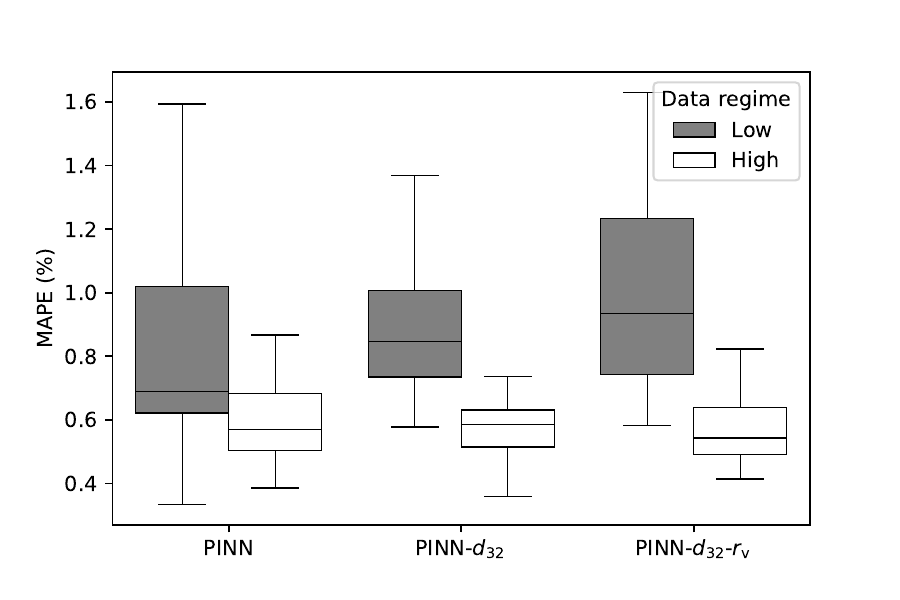}
    \caption{Sedimentation rate $\Dot{V}_s$.}
    \label{fig:0d-settler-qs_wo_NN}
\end{subfigure}
\caption{Test set error for the water height $h_{\textrm{aq}}$ and the sedimentation rate $\Dot{V}_s$ for all PINN models and data regimes. The error metric is the mean absolute percentage error (MAPE).}
\label{fig:test}
\end{figure}

In Figure \ref{fig:0d-settler-haq_wo_NN}, we observe that the prediction accuracy of the water height $h_{\textrm{aq}}$ does not change notably with the addition of  $d_{32}$ and $r_{\textrm{v}}$ as further NN inputs. We note a similar trend for the prediction of the sedimentation rate $\Dot{V}_s$ in Figure \ref{fig:0d-settler-qs_wo_NN}. These findings are not unexpected since the added physical properties play a negligible role in the sub-model for sedimentation rate $\Dot{V}_s$ in the full-order mechanistic model and consequently for the water height $h_{\textrm{aq}}$.

The vanilla ANN performs considerably worse: For the DPZ height $h_{\textrm{DPZ}}$, the mean error of 25 models is 2.06 \% (MAPE) for the low-data regime and 1.78 \% (MAPE) for the high-data regime. For the water phase height $h_{\textrm{aq}}$, the mean error of 25 models is 1.33 \% (MAPE) for the low-data regime and 1.13 \% (MAPE) for the high-data regime. Note that the vanilla ANN cannot estimate the coalescence and sedimentation rates, $\Dot{V}_c$ and $\Dot{V}_s$, as no measurement data were available for training. 

Overall, all PINN models show great generalization capabilities in the low-data regime for the prediction of $h_{\textrm{DPZ}}$ which is a significant performance indicator for separation efficiency, with a maximum value of 0.46 \% for the mean absolute percentage error (MAPE). Moreover, the PINN models can estimate the unmeasured states, for which constitutive equations were assumed to be unknown, with a maximum error value of 8.28 \% for the coalescence rate $\Dot{V}_c$, and with a maximum error value of 1.62 \% for the sedimentation rate $\Dot{V}_s$. As a final remark, we observe that adding $d_{32}$ and $r_{\textrm{v}}$ as inputs to the PINN significantly improves the prediction of $h_{\textrm{DPZ}}$ and the estimation of $\Dot{V}_c$.

\section{Conclusion and Outlook}\label{sec:conclusion} 

This paper investigates the PINN-based dynamic modeling of chemical engineering processes that are characterized by limited physical knowledge and limited data availability. 
Recognizing that certain process states, e.g., reaction rates or coalescence rates, often lack descriptive constitutive equations and cannot 
be measured, we set out to see if PINNs can infer such unmeasured states by leveraging known physical equations and data on measured states. To this end, we conducted numerical studies using two fully-known mechanistic process models and mimicking real-world modeling situations that are characterized by limited physical knowledge and data availability. Specifically, we assumed that certain equations would be unknown and thus unavailable for PINN development and that only a subset of process states would be measurable. 

In both the Van de Vusse continuously stirred tank reactor (CSTR) example and the liquid-liquid separator example, we found that PINN models vastly outperform vanilla NNs of equal size, show superior generalization with respect to different initial states and control actions as well as superior extrapolation capabilities in regions of the state space without training data. 
Importantly, we observed that PINN models indeed may be capable of estimating unmeasured states, even if the corresponding constitutive relations are unknown. We provided a heuristic for when the estimation of such unmeasured states might be successful. Although representing neither a necessary nor a sufficient condition for state estimation, the heuristic is easy to use and can be applied even before data collection is initiated. 

Future work should concern the investigation of implicit DAE models in PINNs and whether the heuristic can be improved based on theory for observability of nonlinear dynamic systems, see, e.g., \cite{lee1967foundations,kou1973observability}. The feed-forward PINNs used in our work could also be compared to physics-informed \emph{recurrent} neural networks that rely on time discretization, see, e.g., \cite{ZHENG2023103005}, or transformer-based PINN architectures, see, e.g., \cite{zhao2023pinnsformer}. Furthermore, PINN modeling and control of actual plant operations should be considered.

We conclude that PINN models with partial physical knowledge constitute a promising alternative to hybrid mechanistic/data-driven models in chemical engineering applications and warrant further investigation by the PSE community due to their potential to estimate states for which neither constitutive equations nor training data are available. Such further investigation should include performance comparisons between PINNs and hybrid models on identical tasks. For instance, for the estimation of immeasurable states $\bm y$ that lack constitutive equations, a DNN predicting measurable differential states $\bm x$ followed by a mechanistic model that uses (i) known balance equations, (ii) the predictions of $\bm x$, and (iii) estimates of $\dot{\bm x}$ obtained through automatic differentiation of the DNN to compute immeasurable algebraic states $\bm y$ would constitute a sequential hybrid model that could be compared to a PINN. Similarly, hybrid models recently proposed by \citet{PahariLatent} and \citet{sitapurehybrid} use DNNs and time-series transformers, respectively, to estimate (spatio-)temporally varying quantities as inputs to mechanistic sub-models.

\section*{Declaration of Competing Interest}
We have no conflict of interest.

\section*{Acknowledgements}
\label{sec:acknowledgements}

This work was funded by the Deutsche Forschungsgemeinschaft (DFG, German Research Foundation) – 466656378 – within the Priority Programme “SPP 2331:Machine Learning in Chemical Engineering”. This work was performed as part of the Helmholtz School for Data Science in Life, Earth and Energy (HDS-LEE). We acknowledge financial support by the Helmholtz Association of German Research Centers 
through program-oriented funding.
We would like to give special thanks to Lukas Polte, Fabian Mausbeck and Lukas Thiel (Aachener Verfahrenstechnik, Fluid Process Engineering, RWTH Aachen University) for their valuable suggestions and dedicated help on the separator model. We would like to give special thanks to Adel Mhamdi (Aachener Verfahrenstechnik, Process Systems Engineering, RWTH Aachen University) for providing valuable insight into the concepts of state estimation and observability.

\section*{Author contributions}

\begin{itemize}
    \item MV developed the PINN-based dynamic models, implemented the PINN model for the Van de Vusse CSTR example, analyzed the results, and wrote the draft of all sections except the system description in Section \ref{sec:Settler} and the separator related parts of the Supplementary Materials. 
    \item MV implemented the PINN model for the liquid-liquid separator example in close collaboration with SZ.
    \item SZ implemented and extended the mechanistic model of the liquid-liquid separator from the literature \citep{Backi2018AFirst-Principles,Backi.2019}, wrote the system description in Section \ref{sec:Settler} and the separator-related parts of the Supplementary Materials. 
    \item SZ and MV jointly analyzed the PINN results for the separator example and wrote Section \ref{sec:settler-results}.
    \item SR implemented the mechanistic model of the Van de Vusse CSTR, developed a preliminary PINN model for the Van de Vusse CSTR, and investigated scaling and dynamic weighting of PINNs in close collaboration with MV.
    \item MD conceptualized and supervised the work with the exception of the derivation of the mechanistic separator model, and provided help and guidance on the methodology.
    \item AM provided conceptual input on the theory, methods, and case studies and provided further supervision.
    \item AJ provided conceptual input on the separator model and provided further supervision.
    \item All authors have reviewed and edited the manuscript.
\end{itemize}

\section*{CRediT authorship contribution statement}

\textbf{Mehmet Velioglu}: Conceptualization, Methodology, Software, Investigation, Writing - original draft, review \& editing, Visualization, Supervision. \noindent \textbf{Song Zhai}: Conceptualization, Methodology, Software, Investigation, Writing - original draft, review \& editing, Visualization. \noindent \textbf{Sophia Rupprecht}:  Methodology, Software, Investigation, Writing - review \& editing. \noindent \textbf{Andreas Jupke}: Conceptualization, Writing - review \& editing, Supervision, Funding acquisition. \noindent \textbf{Alexander Mitsos}: Conceptualization, Methodology, Writing - review \& editing, Supervision.
\noindent \textbf{Manuel Dahmen}: Conceptualization, Methodology, Writing - review \& editing, Supervision, Funding acquisition.

  \renewcommand{\refname}{Bibliography}  
  \bibliography{ms.bib}
  \bibliographystyle{elsarticle-harv}

\end{document}


\doublespacing

  \thispagestyle{firststyle}

  \begin{center}
    \begin{large}
      \textbf{\smtitle}
    \end{large} \\
    \myauthor
  \end{center}

  \vspace{0.5cm}

  \begin{footnotesize}
    \affil
  \end{footnotesize}

  \vspace{0.5cm}

\section*{SM1: Figures of the PINN models used in the numerical examples}
\label{app:SM1} 

\begin{figure}[htbp]
    \centering
\begin{subfigure}{0.6\textwidth}
    \includegraphics[width=\linewidth]{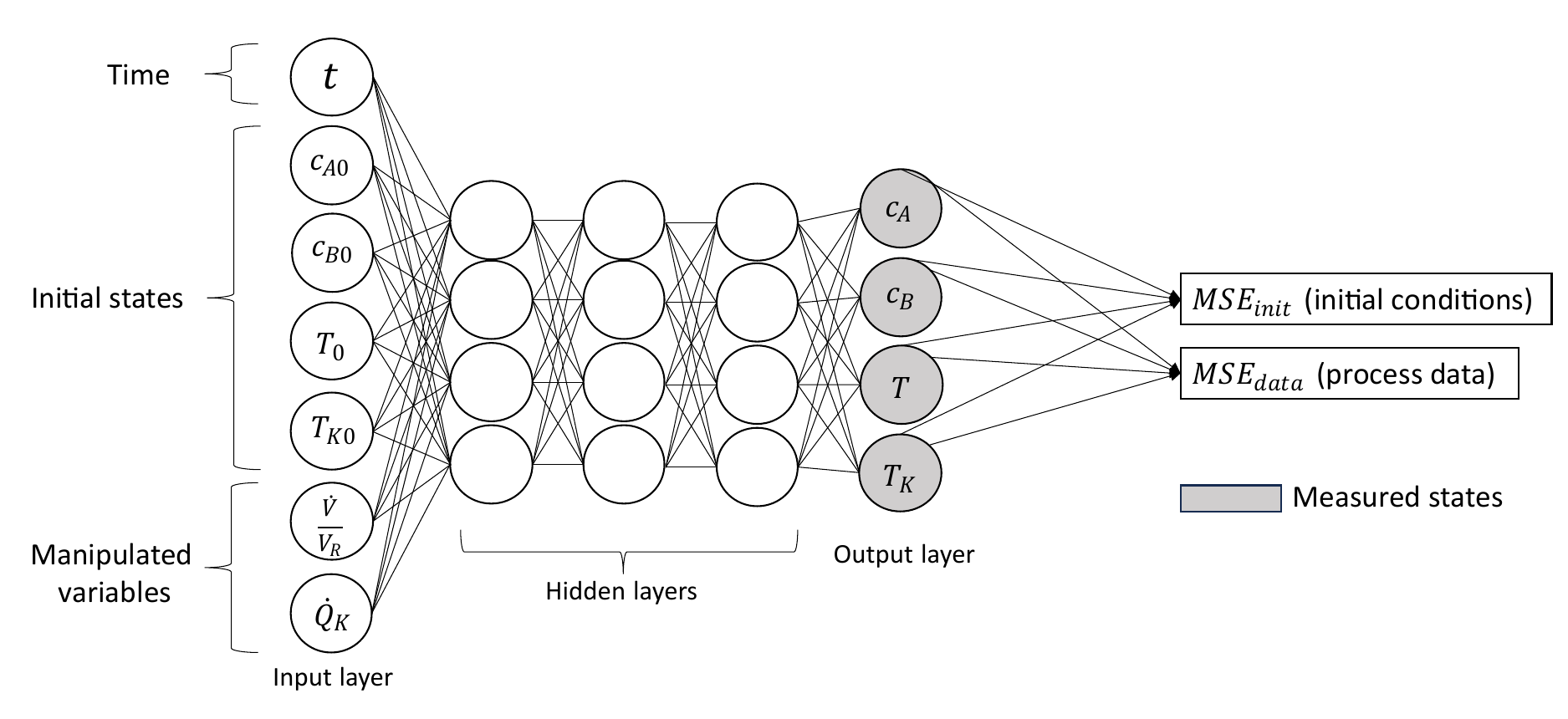}
    \caption{Network schematic of the vanilla ANN}
    \label{fig:vdv-nn}
\end{subfigure}
\vfill
\begin{subfigure}{0.6\textwidth}
    \includegraphics[width=\linewidth]{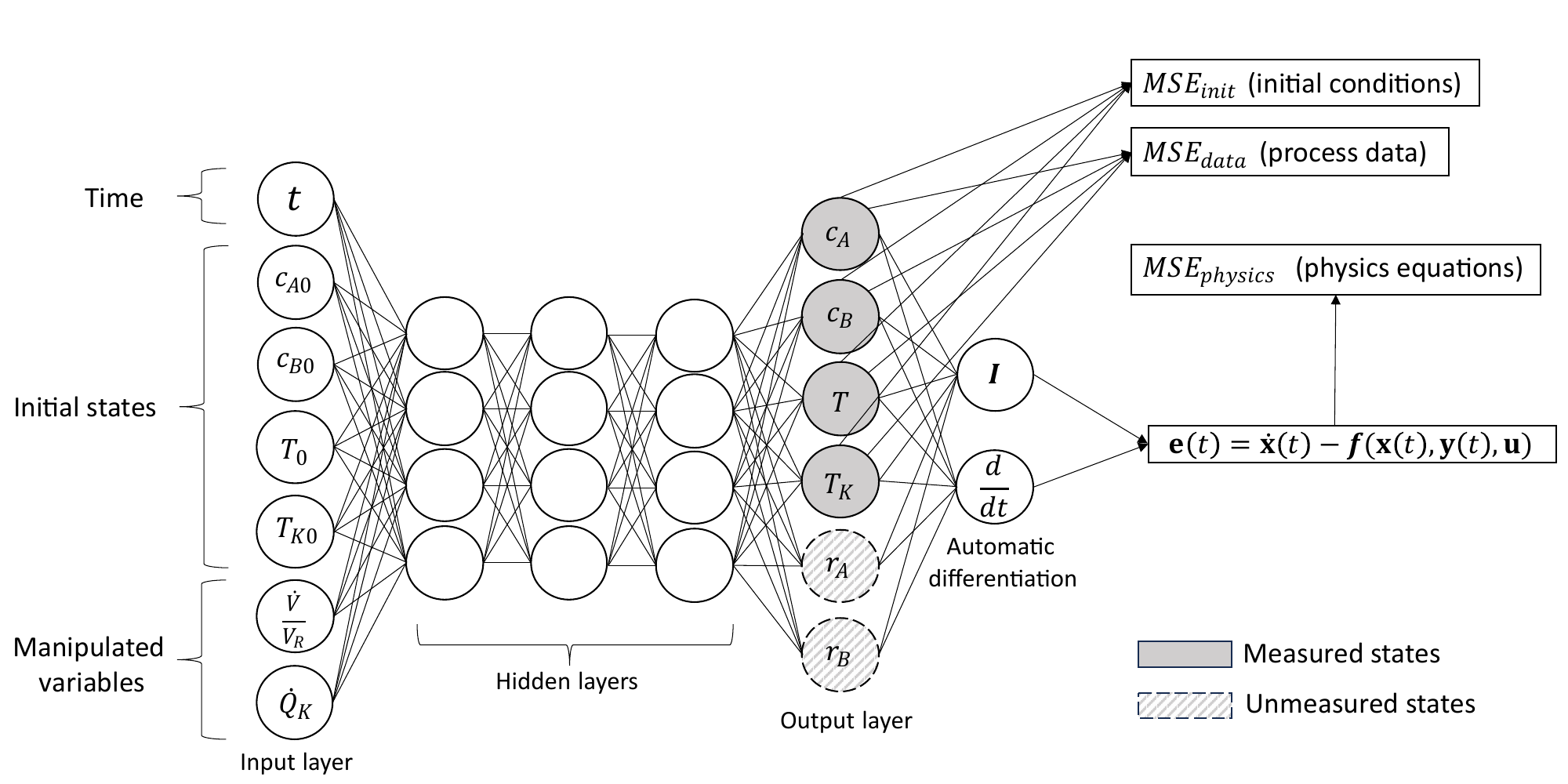}
    \caption{Network schematic of PINN-A}
    \label{fig:vdv-pinn-a}
\end{subfigure}
\vfill
\begin{subfigure}{0.6\textwidth}
    \includegraphics[width=\linewidth]{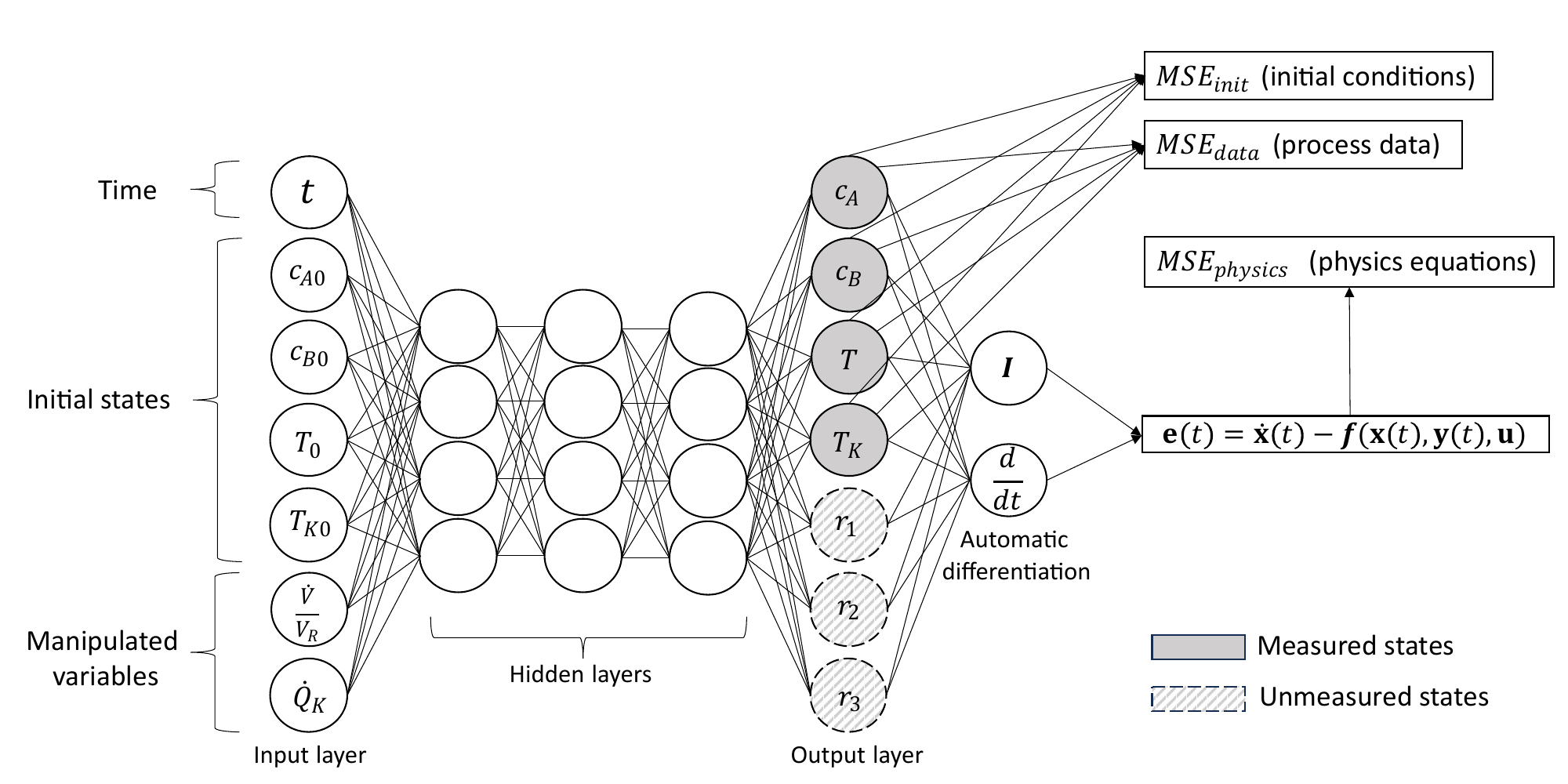}
    \caption{Network schematic of PINN-B}
    \label{fig:vdv-pinn-b}
\end{subfigure}
\vfill
\begin{subfigure}{0.6\textwidth}
    \includegraphics[width=\linewidth]{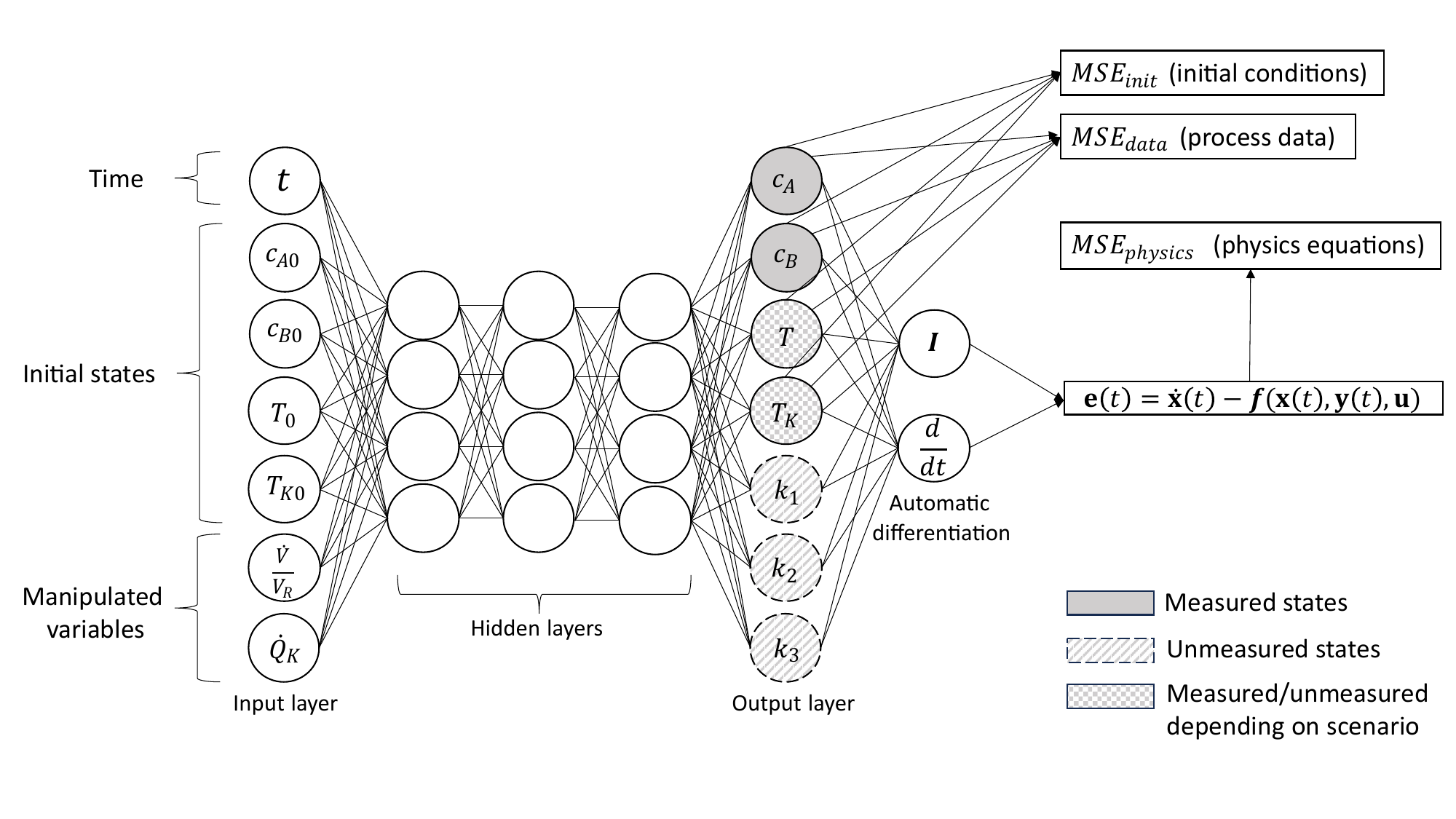}
    \caption{Network schematic of PINN-C}
    \label{fig:vdv-pinn-c}
\end{subfigure}
\caption{Network schematics of models used in the Van de Vusse CSTR example. Note that the figure does not show the actual depth and width of the hidden layers.}
\label{fig:vdv-models}
\end{figure}

The network schematics of the models for the Van de Vusse CSTR are shown in Figure \ref{fig:vdv-models}.

\begin{figure}[htbp]
    \centering
    \includegraphics[scale=0.5]{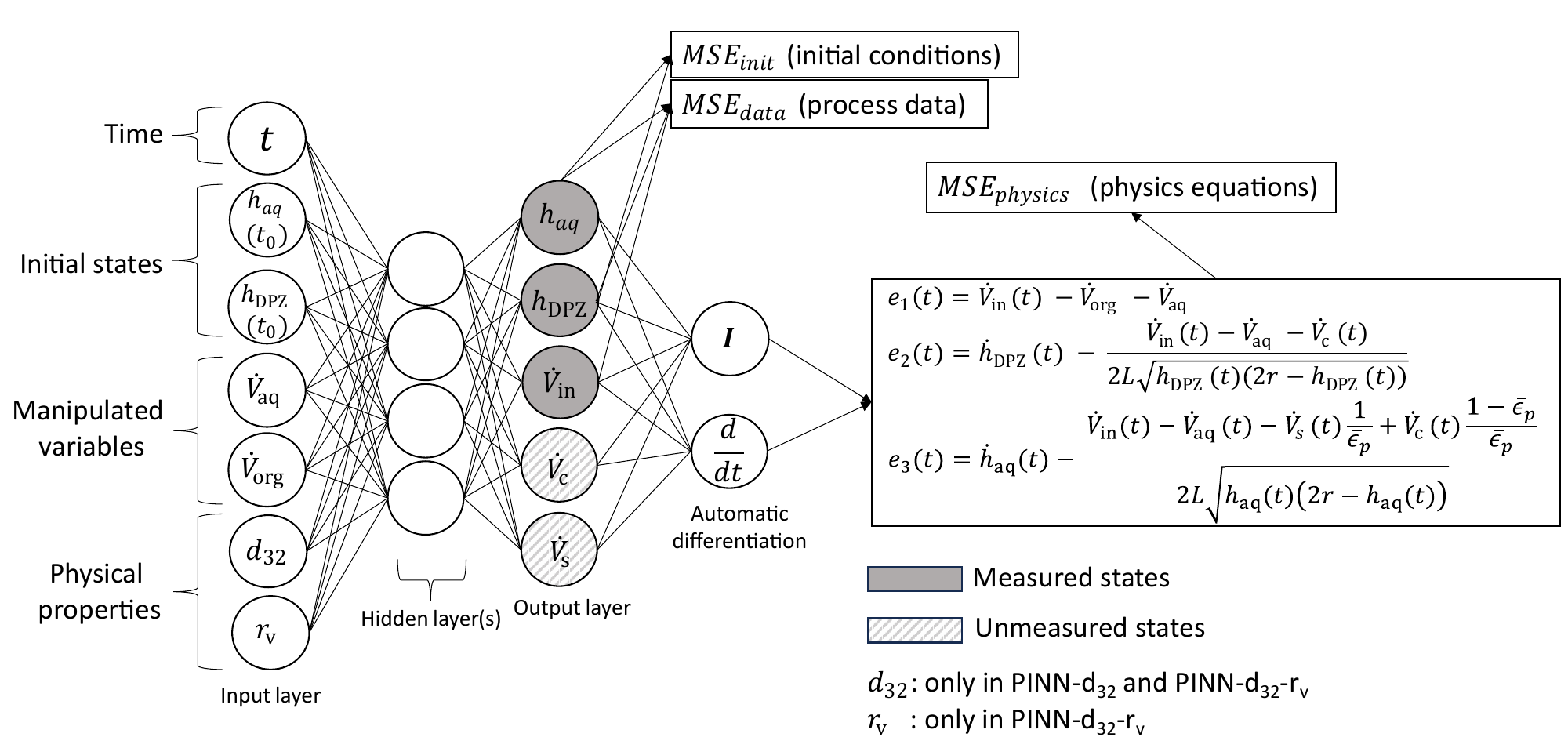}
    \caption{General schematic of the PINN models for the liquid-liquid separator. $\Dot{V}_{\textrm{aq,out}}$ is shown as $\Dot{V}_{\textrm{aq}}$ and $\Dot{V}_{\textrm{org,out}}$ is shown as $\Dot{V}_{\textrm{org}}$. Model "{PINN}" has no physical property as input; model "{PINN-$\textrm{d}_\textrm{32}$}" has the Sauter mean diameter $d_{32}$ as a NN input; model "{PINN-$\textrm{d}_\textrm{32}$-$\textrm{r}_\textrm{v}$}" has both the Sauter mean diameter $d_{32}$ and the coalescence parameter $r_{\textrm{v}}$ as NN inputs. Note that the figure does not show the actual depth and width of the hidden layers.}
    \label{fig:0d-settler-pinn-model}
\end{figure}

The schematic of the PINN models for the liquid-liquid separator is shown in Figure \ref{fig:0d-settler-pinn-model}.

\section*{SM2: Dimensionless variables and equations of the Van de Vusse Reactor}
\label{app:SM2} 

The equations for the Van de Vusse CSTR are given in Equations (3) in the main text. To make the problem dimensionless, we introduce the following dimensionless quantities, inspired by the work of \citet{GAMBOATORRES2000481}: 
\begin{alignat*}{5}
    & \quad t^* = \frac{t}{\tau}, \quad && c_A^* = \frac{c_A}{c_{A,in}}, \quad && c_B^* = \frac{c_B}{c_{A,in}}, \quad && T^* = \frac{T}{T_{in}}, \quad && T_K^* = \frac{T_K}{T_{in}}, \\
    & \left(\frac{\Dot{V}}{V_R}\right)^* = \frac{\Dot{V}}{V_R}\frac{1}{q_f}, \quad && \Dot{Q}_K^* = \frac{\Dot{Q}_K}{\Dot{Q}_{K,f}},  \quad && k_{1}^* = \frac{k_{1}}{k_f}, \quad && k_{2}^* = \frac{k_{2}}{k_f}, \quad && k_3^* = c_{A,in}\frac{k_3}{k_f}
    \label{eq:app-make-dimless}
\end{alignat*}
    
The values of the normalization parameters $\tau$, $q_f$, $\Dot{Q}_{K,f}$ and $k_f$ are given in Table \ref{tab:vdv-dim-param}. For better readability, we introduce the following symbols:
\begin{equation*}
    \qquad P = -\frac{\Delta H_{AB} k_f c_{A,in}}{\rho C_p T_{in}}, \qquad M = \frac{k_w A_R}{\rho C_p V_R}, \qquad L =  \frac{k_w A_R}{m_K C_{pK}}, \qquad R = \frac{\Dot{Q}_{K,f}}{m_K C_{pK} T_{in}}
\end{equation*}

Then, the dimensionless versions of the Equations (3) in the main text become:
\begin{align*}
        \frac{1}{\tau}\left(\dv{c_A^*}{t^*}\right) &= (q_f)\left(\frac{\Dot{V}}{V_R}\right)^*  (1 - c_A^*) - k_f k_1^*(T) c_A^*  - k_f k_3^*(T) {c_A^*}^2, \\
        \frac{1}{\tau}\left(\dv{c_B^*}{t^*}\right)  &= -(q_f)\left(\frac{\Dot{V}}{V_R}\right)^*  c_B^* + k_f k_1^*(T) c_A^*- k_f k_2^*(T) c_B^*, \\
        \begin{split}
        \frac{1}{\tau}\left(\dv{T^*}{t^*}\right)    &= (q_f)\left(\frac{\Dot{V}}{V_R}\right)^* (1 - T^*) + P \left[ k_1^*(T) c_A^* + k_2^*(T) c_B^* \frac{\Delta H_{BC}}{\Delta H_{AB}} + k_3^*(T) {c_A^*}^2\frac{\Delta H_{AD}}{\Delta H_{AB}} \right] \\
        & \qquad + M (T_K^* - T^*),
        \end{split}
        \\
        \frac{1}{\tau}\left(\dv{T_K^*}{t^*}\right)  &= L (T^* - T_K^*) + \Dot{Q}_K^* R
\end{align*}
with
\begin{equation*}
    k_i^*(T^*) = \frac{k_{i0} \exp (\frac{E_i / T_{in}}{T^*})}{k_f}, \qquad i = 1,2,3
\end{equation*}

\begin{table}[h]
\centering
\caption{Parameters for the dimensionless van de Vusse CSTR equations. The values are chosen to bound time and control actions between 0 and 1.}
\begin{tabular}{l | l} 
Symbol & Value\\
\hline
$\tau$ & \SI{60}{\second} \\
$q_f$ & \SI{28.4}{(1\per\hour)} \\
$\Dot{Q}_{K,f}$ & \SI{-2227}{\kilo\joule\per\hour} \\
$k_f$ & \SI{36}{(1\per\hour)} \\
\end{tabular}
\label{tab:vdv-dim-param}
\end{table}

\section*{SM3: Dimensionless variables and equations of the liquid-liquid separator case study}
\label{app:SM3} 

The equations for the liquid-liquid separator are given in Equations (5) in the main text. To make the problem dimensionless, we introduce the following dimensionless quantities: 
\begin{equation*}
    t^* = \frac{t}{\tau}, \quad h_L^* = \frac{h_L}{2r}, \quad h_{DPZ}^* = \frac{h_{DPZ}}{2r}, \quad h_{\textrm{aq}}^* = \frac{h_{\textrm{aq}}}{2r}, \quad \Dot{V}_i^* = \frac{\Dot{V}_i}{q_f}, \\
\end{equation*}    
The values of the normalization parameters above are given in Table \ref{tab:settler-dim-param}.
Then the dimensionless version becomes:
\begin{align*}
        \frac{1}{\tau}\left(\dv{h_L^*}{t^*}\right) &= \frac{q_f}{2r} \frac{\dot{V}_{in}^* - \Dot{V}_{aq,out}^* - \Dot{V}_{org,out}^*}{2L\sqrt{2r h_L^*(2r-2r h_L^*)}}, \\
        \frac{1}{\tau}\left(\dv{h_{DPZ}^*}{t^*}\right)  &= \frac{q_f}{2r} \frac{\dot{V}_{in}^* - \Dot{V}_{aq,out}^* - \Dot{V}_c^* } {2L\sqrt{2r h_{DPZ}^*(2r-2r h_{DPZ}^*)}}\\
        \frac{1}{\tau}\left(\dv{h_{\textrm{aq}}^*}{t^*}\right)  &= \frac{q_f}{2r} \frac{\dot{V}_{in}^* - \Dot{V}_{aq,out}^* - \Dot{V}_s^* \frac{1}{\bar{\epsilon}_p} +  \Dot{V}_c^* \frac{1-\bar{\epsilon}_p}{\bar{\epsilon}_p} }{2L\sqrt{2r h_{\textrm{aq}}^*(2r-2r h_{\textrm{aq}}^*)}}
\end{align*}
\begin{table}[h]
\centering
\caption{Parameters for the dimensionless liquid-liquid separator equations. The values are chosen to bound time, control actions, and states between 0 and 1.}
\begin{tabular}{l | l} 
Symbol & Value\\
\hline
$\tau$ & \SI{20}{\second} \\
$q_f$ & \SI{1e-3}{\cubic\meter\per\second} \\
\end{tabular}
\label{tab:settler-dim-param}
\end{table}

\section*{SM4: Lumped model of a liquid-liquid separator}
\label{app:SM4}

\begin{figure}[h]
    \centering
    \includegraphics[scale=0.49]{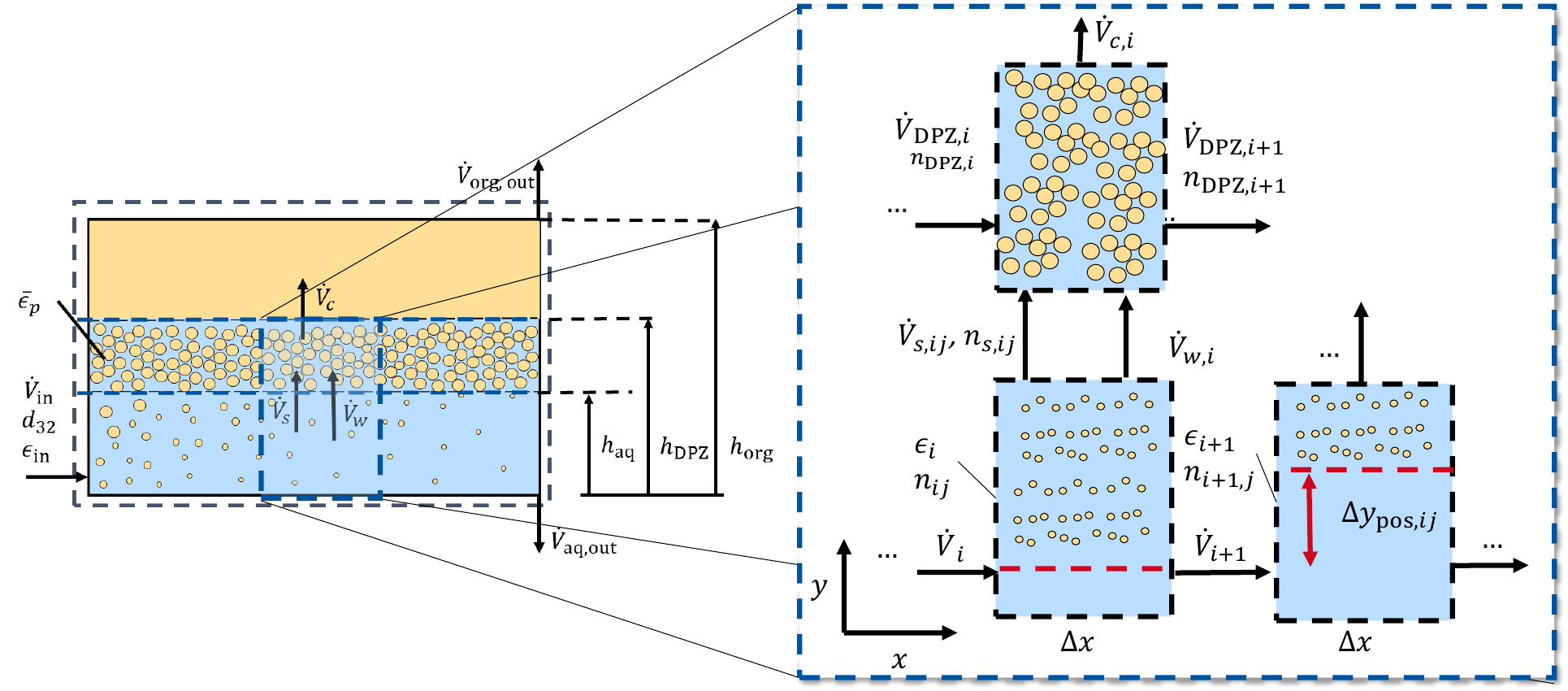}
    \caption{Detailed 1D model of the heavy phase and DPZ at an axial element $i$ and for a droplet class $j$. $\dot{V}_i$ is the convective flow of the aqueous phase, $\epsilon_i$ the hold-up in the aqueous phase, $n_{i,j}$ the number of droplets. $\Delta x$ is the discretization length in axial direction and $\Delta y_{\textrm{pos},ij}$ is the sedimented distance of droplets. $\dot{V}_{s,ij}$, $\dot{V}_{w,i}$ and $\dot{V}_{c,i}$ are the volume flow of sedimenting organic droplets, trapped water, and coalescing organic, respectively. $\dot{V}_{dpz,i}$ and $n_{dpz,i}$ are the convective flows of the DPZ and number distribution in the DPZ.}
    \label{fig:1D_Model}
\end{figure}

The sedimentation and coalescence rate, $\dot{V}_s$ and $\dot{V}_c$, respectively, are determined by lumping a detailed 1D model with discretized drop population balance shown in Fig.~\ref{fig:1D_Model}.
The dynamic liquid-liquid separator model shown below is based on the work of \cite{Backi2018AFirst-Principles, Backi.2019}. We included extensions for swarm sedimentation in the aqueous phase, convection terms for the drop size distribution (DSD) in the dense-packed zone (DPZ) analogously to \cite{Backi2018AFirst-Principles}, and a state-of-the-art coalescence model \citep{Henschke1995DimensionierungAbsetzversuche}. The chosen swarm model \citep{Mersmann.1980} was also used to model liquid-liquid columns \citep{Kampwerth2020TowardsModel} and takes the form of Stokes' law \citep{Stokes1850OnIII} for diminishing hold-ups. Stokes' law was experimentally confirmed to model the outlet hold-up of liquid-liquid separator accurately \citep{Ye2023ImpactEfficiency}. 

Three liquid phases are modeled, aqueous, DPZ, and organic. The organic phase is dispersed, and the continuous organic phase is free from aqueous droplets. The lumped model assumes given constant heights for each phase, an instantaneous development of sedimenting droplets, and a DPZ starting from the entrance with given boundary conditions from the 0D settler model (cf. Section 4 in the main text). In Addition, no coalescence in the aqueous phase is assumed.

In the following, expressions for the water volume flow rate $\dot{V}_{w}$, sedimentation rate $\dot{V}_s$, and coalescence rate $\dot{V}_c$ are derived.
The water volume flow $\dot{V}_w$ from the aqueous phase can be written as a function of the coalescence and sedimentation rates resulting from a volume balance for the DPZ by assuming a constant hold-up $\bar{\epsilon}_p$ in the DPZ:
\begin{align}
    \dv{V_{\textrm{DPZ}}}{t} &= \dot{V}_s + \dot{V}_w - \dot{V}_c \\
    \bar{\epsilon}_p \dv{ V_{\textrm{DPZ}}}{t} &= \dot{V}_s - \dot{V}_c \\
    \Rightarrow \dot{V}_w &= (\dot{V}_s - \dot{V}_c) \frac{1-\bar{\epsilon}_p}{\bar{\epsilon}_p}\label{eq:V_w}
\end{align}

The sedimentation rates $\dot{V}_s$ result from droplets sedimenting with the swarm sedimentation velocity to the interface and moving in a plug flow in the horizontal direction.
The droplets are assumed to move with the same velocity as the heavy phase in $x$-direction and are homogeneously distributed in the heavy phase at the entrance.
The sedimentation rate is calculated as
\begin{align*}
    \dot{V}_s &= \sum_i^{N_s} \dot{V}_{s,i}\quad, \\ 
    \dot{V}_{s,i} &= \dot{V}_i \sum_j^{N_d} \frac{n_{s,i,j}}{n_i}\quad,
\end{align*}
where $\dot{V}_{s,i}$ and $\dot{V}_i$ is the volume flow by sedimentation and convective volume flow in segment $i$, $n_{s, i,j}$ is the number of droplets in class $j$ and in axial segment $i$ reaching the DPZ, and $n_i$ is the total number of droplets in segment $i$. $N_s$ and $N_d$ are the number of axial discretization elements and drop diameter classes, respectively. The convective volume flow $\dot{V}_i$ entering segment $i$ is calculated starting from the entrance using the following equations:
\begin{align*}
    \dot{V}_{i+1} &= \dot{V}_{i} - \dot{V}_{s,i} - \dot{V}_{w,i} \qquad \mbox{for} \quad i=0,...,N_s\quad, \\
    \dot{V}_{i=0} &= \dot{V}_\textrm{in}
\end{align*}
Similarly, the total number of drops $n_{i}$ remaining in a segment $i$ is determined as
\begin{align*}
    n_{i} &= \sum_j^{N_d} n_{i,j}\quad, \\
    n_{i+1,j} &= n_{i,j} - n_{s,i,j} \qquad \mbox{for} \quad i=0,...,N_s \quad \mbox{and} \quad j=1,...,N_d\quad,\\
    n_{i=0,j} &= n_{0,j}\quad,
\end{align*}
where $n_{0,j}$ results from the droplet number distribution at the inlet. The Sauter mean diameter at the inlet and number distribution are related by assuming a self-similar volume-based log-normal drop size distribution with a normalized standard deviation of $\sigma/d_{32}=0.32$ \citep{Kraume2004InfluenceDispersions, Ye2023EffectFractions}. \\
The last missing part is the determination of sedimented droplets $n_{s,i,j}$, which is calculated as
\begin{align*}
    n_{s,i,j} &= n_{i,j} \frac{\tau_{x,i} v_{s,j,i}}{h_{\textrm{aq}} - y_{i,j}} \qquad & \mbox{if} \quad \tau_{x,i} < \tau_{y,i,j}\quad, \\
    n_{s,i,j} &= n_{i,j} \qquad & \mbox{if} \quad \tau_{x,i} \geq \tau_{y,i,j}\quad,
\end{align*}
where $\tau_{x,i}$ and $\tau_{y,i,j}$ are the residence time in x- and y-direction. $v_{s,j,i}$ is the swarm sedimentation of droplet class $j$ in segment $i$, $h_{\textrm{aq}}$ is the height of the aqueous phase, and $y_{i,j}$ is the vertical position of droplet class $j$ in segment $i$. The residence times are determined as
\begin{align}
    \tau_{x,i} &= \frac{V_{\textrm{aq}}/N_s}{\dot{V}_i}\quad, \\
    \tau_{y,i,j} &= \frac{h_{\textrm{aq}} - y_{i,j}}{v_{s,i,j}}\quad, \\
    v_{s,i,j} &= \frac{gd_j^2\Delta\rho}{18 \eta_c} (1-\epsilon_i)^{(n-1)}\quad, \label{eq:swarm_sed}
\end{align}
where $v_{s,i,j}$ is the swarm sedimentation velocity calculated with the swarm exponent ($n=2$) \citep{Mersmann.1980,Kampwerth2020TowardsModel}, $g$ the gravitational constant, $\Delta\rho$ the density difference between the aqueous and organic phase, $\eta_c$ the viscosity of the continuous phase, and $\epsilon_i$ the hold-up in the aqueous phase at segment $i$. For hold-ups approaching \SI{0}{}, Equation \eqref{eq:swarm_sed} takes the form of Stokes' law \citep{Stokes1850OnIII} that was experimentally confirmed to model the outlet hold-up accurately \citep{Ye2023ImpactEfficiency}.
$h_{\textrm{aq}}$ is the height of the aqueous phase and a function of the volume of the aqueous phase, radius $R$, and length $L$ of the separator. The geometric equations are given as follows:
\begin{align*}
    V_{\textrm{aq}} &= A_x(h_{\textrm{aq}})L \quad,\\
    A_x(h) &= R^2 \arccos(1-h/R) - (R-h) \sqrt{2Rh - h^2} 
\end{align*}
The hold-up and vertical position are calculated similarly to the convective volume flow starting from the entrance using the following equations:
\begin{align*}
    \epsilon_{i+1} &= \frac{\epsilon_i \dot{V}_i - \dot{V}_{s,i}}{\dot{V}_{s,i+1}} \qquad & \mbox{for} \quad i=0,...,N_s \quad,\\
    \epsilon_{i=0} &= \epsilon_{\textrm{in}}\quad, \\
    y_{i+1,j} &= y_{i,j} + v_{s,i,j} \min{(\tau_{x,i}, \tau_{y,i,j})} \quad & \textrm{for} \quad i=0,...,N_s ; j=1,...,N_d\quad, \\
    y_{i=0, j} &= 0 \quad & \mbox{for} \quad j=1,...,N_d
\end{align*}
Thus, the sedimentation rate can be calculated from a given height and the boundary conditions. The coalescence rate influences the sedimentation rate indirectly, as the water volume flow rate is a function of the coalescence rate (see Eq.~\eqref{eq:V_w}). Therefore, the coalescence rate influences the convective flow and the residence time for droplets to sediment. 

The coalescence rate $\dot{V}_{c}$ is a function of the Sauter mean diameter, height of the DPZ, and physical properties. The coalescence rate is adopted from \citep{Henschke1995DimensionierungAbsetzversuche} and is calculated as
\begin{align*}
    \dot{V}_c &= \sum_i^{N_s} \dot{V}_{c,i} = \sum_i^{N_s} \frac{2 A_y d_{32,\textrm{DPZ},i}}{3 \tau_{di,i}}\quad, \\
    A_y &= 2\Delta x \sqrt{2Rh_{\textrm{DPZ}} - h_{\textrm{DPZ}}^2}\quad,
\end{align*}
where $A_y$ is the area in the y-direction between the organic phase and DPZ and is a function of the height of the DPZ, $h_{\textrm{DPZ}}$. $d_{32,\textrm{DPZ},i}$ is the Sauter mean diameter in the DPZ at segment $i$, and $\tau_{di,i}$ is the coalescence time at segment $i$. The coalescence time depends on physical properties and the height of the DPZ \citep{Henschke1995DimensionierungAbsetzversuche}, i.e.,
\begin{align*}
    \tau_{di,i} &= \frac{(6\pi)^{7/6} \eta_c r_{\textrm{a}}^{7/3}}{4 \sigma^{5/6} H_c^{1/6} r_{\textrm{f,i}} r_{\textrm{v}}}\quad,  \\
    r_{\textrm{f,i}} &= 0.5239\, d_{32,\textrm{DPZ},i} \sqrt{1-\frac{4.7}{La_{\textrm{mod},i} + 4.7}}\quad, \\
    r_{\textrm{a}} &= 0.5\, d_{32,\textrm{DPZ},i} \Bigg( 1 - \sqrt{1-\frac{4.7}{La_{\textrm{mod},i} + 4.7}} \Bigg)\quad, \\
    La_{\textrm{mod},i} &= \bigg(\frac{\Delta\rho g }{\sigma} \bigg)^{0.6} (h_{\textrm{DPZ}} - h_{\textrm{aq}})^{0.2} d_{32,\textrm{DPZ},i}\quad,
\end{align*}
where $\sigma$ is the interfacial tension, $H_c$ is the Hamaker constant fixed to \SI{10E-20}{\newton\meter}, $r_{\textrm{f,i}}$ and $r_{\textrm{a}}$ are radii resulting from deformed droplets between the drop-interface. The drop deformation is characterized by the modified Laplace number $La_{\textrm{mod}}$ representing the ratio between hydrostatic pressure and interfacial tension. $r_{\textrm{v}}$ is the coalescence parameter specific to a liquid-liquid system describing the coalescence affinity and can be determined by batch settling experiment with the liquid-liquid system. 
The last missing variable is $d_{32,\textrm{DPZ},i}$ that is a function of the number distribution of the drops $n_{\textrm{DPZ},i,j}$ in the segment $i$, previous segment $i-1$, and the convective flow $\dot{V}_{\textrm{DPZ},i}$ in the DPZ. The number distribution of drops in the segment $i$ is calculated from the sedimenting drops $n_{s,i}$ as:
\begin{align*}
    n_{\textrm{DPZ},i,j} &= n_{\textrm{DPZ},i-1,j} + n_{s,i,j}   \quad &\mbox{if} \quad \dot{V}_{\textrm{DPZ},i} > 0\quad, \\
    n_{\textrm{DPZ},i,j} &= n_{s,i,j} \quad &\mbox{if} \quad \dot{V}_{\textrm{DPZ},i} = 0\quad, \\
    \dot{V}_{\textrm{DPZ},i+1} &= \max \bigg(\dot{V}_{\textrm{DPZ},i} + \frac{\dot{V}_{s,i} - \dot{V}_{c,i}}{\bar{\epsilon}_p}, 0 \bigg)\quad,\\
    d_{32,\textrm{DPZ},i} &= \frac{\sum_j n_{\textrm{DPZ},i,j} d_j^3 }{\sum_j n_{\textrm{DPZ},i,j} d_j^2} 
\end{align*}
The convective flow in the DPZ results from a volume balance and is ensured by the maximum function to be nonnegative. In case of a greater coalescence rate than the sum of convective flow and sedimentation rate, the Sauter mean diameter consists only of sedimenting droplets. 

The mechanistic model is solved by 200 discretization elements for the separator length and 50 discretization elements for the drop size distribution. The number of discretization points was determined by sensitivity studies. Model parameter, physical properties and geometry data are listed in Table~\ref{tab:separator-param}.

\begin{table}[h]
\centering
\ra{1.3}
\caption{Parameters for the liquid-liquid separator with n-butyl acetate dispersed in water.}
\begin{tabular}{l | l| l | l} 
Parameter & Symbol & Value & Source\\
\hline
Radius of separator & $R$ & $\SI{0.1}{\meter}$ & own lab\\
Length of separator & $L$ & $\SI{1.8}{\meter}$ & own lab\\
Gravity constant & $g$ & \SI{9.81}{\meter\per\square\second} & - \\
Density difference & $\Delta\rho$ & \SI{115}{\kilogram\per\cubic\meter} & \citep{Henschke1995DimensionierungAbsetzversuche}\\
Viscosity of organic phase & $\eta_\textrm{org}$ & \SI{0.775}{\milli\pascal\second} & \citep{Henschke1995DimensionierungAbsetzversuche} \\
Viscosity of aqueous phase & $\eta_\textrm{aq}$ & \SI{1.012}{\milli\pascal\second} & \citep{Henschke1995DimensionierungAbsetzversuche} \\
Interfacial tension & $\sigma$ & \SI{0.013}{\newton\per\meter} & \citep{Henschke1995DimensionierungAbsetzversuche} \\
Coalescence parameter & $r_\textrm{v}$ & \SI{0.0383}{} & \citep{Henschke1995DimensionierungAbsetzversuche} \\
Hamacker constant & $H_c$ & \SI{1e-20}{\newton\meter} & \citep{Henschke1995DimensionierungAbsetzversuche} \\
Hold-up in DPZ & $\bar{\epsilon}_p$ & \SI{0.9}{} & \citep{Henschke1995DimensionierungAbsetzversuche} \\
\end{tabular}
\label{tab:separator-param}
\end{table}
\section*{SM5: Counter-example showing that the heuristic is not a necessary condition for the estimation of states}

We present a counter-example showing that our heuristic (Section 2.3 in the main manuscript) is not a necessary condition for the estimation of unmeasured states. Consider the following ordinary differential equation (ODE) system:
\begin{align}
        \dot{x}^m_1(t)&=x^m_1(t)x^u_2(t)+x^u_3(t), \label{ce-1}\\ 
        \dot{x}^u_2(t)&=0, \label{ce-2}\\ 
        \dot{x}^u_3(t)&=0 \label{ce-3} 
\end{align}    
Here, $x^m_1$, $x^u_2$ and $x^u_3$ denote the differential states and $t$ denotes time. Note that this ODE system is a complete system, as it has three equations and three variables. We thus consider the extreme case where we can integrate a full physical model into the PINN. In the following, we assume that $x^m_1$ is measured while $x^u_2$ and $x^u_3$ are unmeasured. 

Table \ref{tab:si-ce} shows the incidence matrix for the counter-example given by Equations \eqref{ce-1}-\eqref{ce-3}. Obviously, the incidence matrix does not have full-column rank, suggesting that state estimation would not work. 

\begin{table}[h]
\centering
\caption{Incidence matrix of the counter-example represented by Equations \eqref{ce-1}-\eqref{ce-3}. If an unmeasured state appears in an equation, it is marked with a cross. The matrix does \emph{not} have full-column rank.}
\label{tab:si-ce}
\begin{tabular}{ l|c|c|c }
$[\bm f, \bm g]\downarrow \quad [\Vx^u, \Vy^u]\rightarrow$          & $x^u_2$ & $x^u_3$ \\ \hline
Eqn. \eqref{ce-1} & $\times$  &  $\times$    \\ \hline
Eqn. \eqref{ce-2} &   &     \\ \hline
Eqn. \eqref{ce-3} &   &   
\end{tabular}
\end{table}

We will show that the initial values $x^u_{2,0} = x^u_2(t=0)$ and $x^u_{3,0} = x^u_3(t=0)$ can be determined using measurement data $x^m_1(t_j)$ that provide trajectory information about $x^m_1$. If the initial states $x^u_{2,0}$ and $x^u_{3,0}$ can be uniquely determined from such measurement data $x^m_1(t_j)$, the states $x^u_2$ and $x^u_3$ are said to be observable \citep{kalman1960general,lee1967foundations}. 

The analytical solution to the ODE system reads:
\begin{align*}
        x^m_1(t)&=\left(x^m_{1,0} + \frac{x^u_{3,0}}{x^u_{2,0}}\right)e^{(x^u_{2,0} t)} - \frac{x^u_{3,0}}{x^u_{2,0}},\\
        x^u_2(t)&=x^u_{2,0},\\
        x^u_3(t)&=x^u_{3,0}
\end{align*}
\label{counter-example-analytical}
Here, $x^m_{1,0} = x^m_1(t=0)$ denotes the initial value for the differential state $x^m_1$. Since we have measurement data $x^m_1(t_j)$, we assume that $x^m_{1,0}$ is known. 
A key concept of observability analysis is the exploitation of derivative information, see, e.g., \cite{lee1967foundations,kou1973observability}. Accordingly, we consider the first and second order derivatives of the analytical solution for the measured state $x^m_1$, i.e., 
\begin{align}
    \dot{x}^m_1(t) &=\left(x^m_{1,0}x^u_{2,0} + x^u_{3,0}\right)e^{(x^u_{2,0} t)},  \label{eq:31} \\
    \ddot{x}^m_1(t) &=\left(x^m_{1,0}{x^u_{2,0}}^2 + x^u_{3,0}x^u_{2,0}\right)e^{(x^u_{2,0} t)}  \label{eq:32}.
\end{align}
Since the PINN is given many samples $x^m_1(t_j)$ for different $t_j$ during training, it is provided with trajectory data on $x^m_1$, which, in principle, allows deriving derivative information. Thus, we assume the derivatives of $x^m_1$ to be known. 
With the assumption of known derivatives $\dot{x}^m_1(t)$ and $\ddot{x}^m_1(t)$, Equations \eqref{eq:31} and \eqref{eq:32} can be solved for the unknown initial states $x^u_{2,0}$ and $x^u_{3,0}$: 
\begin{align*}
 x^u_{2,0} &= \frac{\ddot{x}^m_1(t)}{\dot{x}^m_1(t)}, \\
 x^u_{3,0} &= \frac{\dot{x}^m_1(t)}{e^{\left(\frac{\ddot{x}^m_1(t)}{\dot{x}^m_1(t)} t\right)}} - x^m_{1,0} \frac{\ddot{x}^m_1(t)}{\dot{x}^m_1(t)} 
\end{align*}
Thus, the states $x^u_2$ and $x^u_3$ are observable if $\dot{x}^m_1(t) \neq 0$, although the incidence matrix suggests that state estimation should not work. 

We provide empirical evidence showing that a PINN for the counter-example is indeed capable of estimating the unmeasured states $x^u_2$ and $x^u_3$. 
To this end, we implement a corresponding PINN model that takes $t$ as input and has the outputs $x^m_1(t)$, $x^u_2(t)$ and $x^u_3(t)$. We omit an input $x^m_{1,0}$, since we test the PINN for a single choice of initial values. We provide synthetic measurement data, i.e., samples $x^m_1(t_j)$, as training data, and integrate Equations \eqref{ce-1}-\eqref{ce-3} as physics knowledge into the PINN. Specifically, we create training data using the explicit Runge-Kutta method of order 5, utilizing \textit{solve\_ivp} solver from \textit{scipy.integrate} module in Python \citep{Virtanen2020SciPyPython,Dormand1980AFormulae} and using $x^m_{1,0}=1$, $x^u_{2,0}=1$ and $x^u_{3,0}=2$. The time domain is chosen as $t \in [0,1]$.

\begin{figure}[h]
    \centering
    \includegraphics[scale=0.8]{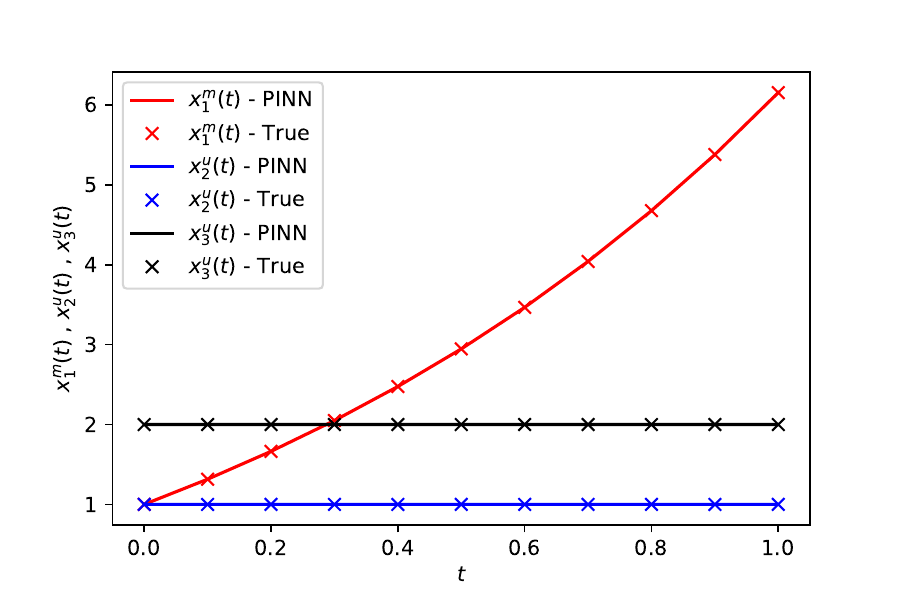}
    \caption{Predicted trajectory of $x^m_1(t)$ and estimated trajectories of $x^u_2(t)$ and $x^u_3(t)$ by the PINN (lines), along with the ground truth trajectory data obtained from the ODE solver (markers).}
    \label{fig:counterexample_pinn}
\end{figure}

As can be seen from Figure \ref{fig:counterexample_pinn}, the PINN correctly estimates $x^u_2(t)$ and $x^u_3(t)$. This finding supports the results obtained from the observability analysis and clarifies that the heuristic based on the incidence matrix does not constitute a necessary condition for state estimation.  

\section*{SM6: Counter-example showing that the heuristic is not a sufficient condition for the estimation of states}

The following counter-example demonstrates that our heuristic (Section 2.3 in the main manuscript) is not a sufficient condition for the estimation of states:
\begin{align}
\dot{x}^u_1 &= x^u_1 + x^u_2 + y^m \label{ce2-1} \\
\dot{x}^u_1 &= \left( (x^u_1)^2 + (x^u_2)^2 + (y^m)^2 + 2x^u_1x^u_2 + 2x^u_1y^m + 2x^u_2y^m \right)^\frac{1}{2} \label{ce2-2}
\end{align}
Here, $y^m$ is the measured state, whereas the differential states $x^u_1$ and $x^u_2$ are unmeasured. 

It is easy to see that Equations \eqref{ce2-1} and \eqref{ce2-2} are dependent:
\begin{align*}
\dot{x}^u_1 &= \left( (x^u_1)^2 + (x^u_2)^2 + (y^m)^2 + 2x^u_1x^u_2 + 2x^u_1y^m + 2x^u_2y^m \right)^\frac{1}{2} \\
&= \left( \left( x^u_1 + x^u_2 + y^m \right) \left( x^u_1 + x^u_2 + y^m \right)\right)^\frac{1}{2} \\
&= x^u_1 + x^u_2 + y^m
\end{align*}
As the ODE system has infinitely many solutions, state estimation is impossible, although the incidence matrix has full-column rank (see Table \ref{tab:si-ce2}). 

\begin{table}[h]
\centering
\caption{Incidence matrix of the counter-example represented by Equations \eqref{ce2-1}-\eqref{ce2-2}. If an unmeasured state appears in an equation, it is marked with a cross. The matrix does have full-column rank.}
\label{tab:si-ce2}
\begin{tabular}{ l|c|c|c }
 $[\bm f, \bm g]\downarrow \quad [\Vx^u, \Vy^u]\rightarrow$         & $x^u_1$ & $x^u_2$ \\ \hline
Eqn. \eqref{ce2-1} & $\otimes$  &  $\times$    \\ \hline
Eqn. \eqref{ce2-2} &  $\times$ & $\otimes$    \\ 
\end{tabular}
\end{table}

\renewcommand{\refname}{Bibliography}  
\bibliography{supplement.bib}
\bibliographystyle{elsarticle-harv}